\documentclass[sn-apa]{sn-jnl}% APA Reference Style 
\usepackage{graphicx}%
\usepackage{multirow}%
\usepackage{amsmath,amssymb,amsfonts}%
\usepackage{amsthm}%
\usepackage{mathrsfs}%
\usepackage[title]{appendix}%
\usepackage{xcolor}%
\usepackage{textcomp}%
\usepackage{manyfoot}%
\usepackage{booktabs}%
\usepackage{algorithm}%
\usepackage{algorithmicx}%
\usepackage{algpseudocode}%
\usepackage{listings}%
\usepackage{array}

\usepackage{subcaption}
\usepackage{multirow}

\usepackage{dcolumn}
\newcolumntype{d}[1]{D{.}{.}{#1}}

\begin{document}

\title[Occlusion-Based Object Transportation Around Obstacles With a Swarm of Miniature Robots]{Occlusion-Based Object Transportation Around Obstacles With a Swarm of Miniature Robots}

%%       \dgr{MSc, PhD}}\email{iauthor@gmail.com}
%%=============================================================%%

\author[1,2]{\fnm{Breno} \sur{Cunha Queiroz}}\email{brenocqueiroz@usp.br}
\equalcont{These authors contributed equally to this work.}

\author*[1]{\fnm{Daniel} \sur{MacRae}}\email{d.macrae@student.rug.nl}
\equalcont{These authors contributed equally to this work.}

\affil[1]{\orgdiv{Faculty of Science and Engineering}, \orgname{Rijksuniversiteit Groningen}, \orgaddress{\city{Groningen}, \country{Netherlands}}}

\affil[2]{\orgdiv{Institute of Mathematics and Computer Sciences}, \orgname{University of São Paulo}, \orgaddress{\city{São Carlos}, \state{São Paulo}, \country{Brazil}}}


\abstract{Swarm robotics utilises decentralised self-organising systems to form complex collective behaviours built from the bottom-up using individuals that have limited capabilities. Previous work has shown that simple occlusion-based strategies can be effective in using swarm robotics for the task of transporting objects to a goal position. However, this strategy requires a clear line-of-sight between the object and the goal. In this paper, we extend this strategy by allowing robots to form sub-goals; enabling any member of the swarm to establish a wider range of visibility of the goal, ultimately forming a chain of sub-goals between the object and the goal position. We do so while preserving the fully decentralised and communication-free nature of the original strategy, while maintaining performance in object-free scenarios. In five sets of simulated experiments, we demonstrate the generalisability of our proposed strategy. Our finite-state machine allows a sufficiently large swarm to transport objects around obstacles that block the goal. The method is robust to varying starting positions and can handle both concave and convex shapes.}




\maketitle

\section{Introduction}\label{sec1}
In swarm robotics, multiple small robots collectively execute a task. These robots are designed to be simple and inexpensive to construct; thereby possessing limited sensing and communication abilities. One task that such robots are applied to is collaborative object manipulation, with a frequently studied example being the box-pushing task \citep{Bayindir2016}. This task calls for robots to move boxes, located in some environment or arena, to a specific target or goal location. The boxes are typically such that they cannot be moved by just a single robot, but require multiple robots to push the object by applying force in the same direction.

Autonomous multi-robot systems capable of cooperative object transportation have the potential to be extremely beneficial in a wide variety of applications, such as waste retrieval and de-mining \citep{Tuci2018}, or operations carried out in situations impractical or challenging for humans, such as space or in deep-sea environments \citep{Huntsberger2000, Parker2006, Woern2006, Farivarnejad2021}. Centralised multi-robot systems have proven effective in warehouses and distribution centres when performing storage tasks in highly controlled environments \citep{Roodbergen2009}, with potential users perceiving the usefulness of swarm robots in facilitating efficient storage, sorting or inventory checking abilities \citep{Carrillo-Zapata2020}. Further studies examine using multiple aerial robots to move heavy objects using cables \citep{Michael2011, Bernard2011}, while cooperative transportation in micro-scale applications, such as molecular delivery to targeted cells, minimally invasive surgery, and tissue engineering \citep{Hu2011, Shahrokhi2016, Rahman2017}, are also gaining traction.

As seen in the next section, a wide variety of approaches have been taken to try and solve the task of object transportation with swarm robots. One novel strategy proposed by Chen et al. (\citeyear{Chen2013, Chen2015}) utilises occlusion to organise the swarm of robots to complete the task of transporting an object, by instructing the robots to push the object whenever it occludes the direct line-of-sight towards the goal. The use of multiple identical robots, each equipped with relatively limited task-solving ability and knowledge about the environment, through their interactions exhibit a behaviour that is complex enough to solve the object transportation task. While this fully decentralised, communication-free, and scalable approach works well in simple environments, \cite{Chen2015} note that this strategy is not effective in complex environments where an obstacle may impede the ability to perceive the goal from positions immediately surrounding the object. While \cite{Chen2015} suggest a method where the robots rely on a human-controlled mobile goal, in this paper we propose an adaptation to this occlusion-based strategy that enables the robots to transport objects around obstacles fully autonomously by adding a behaviour that allows the robots to decide to turn into sub-goals themselves. Our adaptation maintains the fully decentralised, communication-free, vision-based nature of their original strategy. 

The paper is organised as follows. In Section \ref{sec:related_works} we discuss related works in the field of object transportation with swarm robotics. Section \ref{sec:methodology} describes the problem statement, introduces our occlusion-based state machine for the object transportation task, and establishes the conditions used in our simulation experiments. The following five sections each present a different set of experiments. In Section \ref{sec:Obstacle-Free_map_Experiments} we compare this proposed state machine against the original occlusion-based strategy \citep{Chen2015} in an environment where no obstacles are present. Section \ref{sec:Obstacle_map_Experiments} presents a set of experiments performed in environments where at least one obstacle blocks the line of sight between the object and the goal. Section \ref{sec:Different_Shape_Experiments} studies the robustness of our proposed strategy when using differently shaped objects in environments both with and without obstacles. In Section \ref{sec:Starting_Position_Experiments}, we examine the effect of the robots' starting position on their ability to efficiently complete the object transportation task by comparing three different robot position initialisation methods. In Section \ref{sec:Teleop_Experiment}, we compare our strategy to a similar method, but where the sub-goal robot is teleoperated \citep{Chen2015}. This reference strategy departs from our assumption that the environment is not known beforehand, but acts as a heuristic for a more idealised scenario with which we can analyse the nuances of our method in unknown environments. Section \ref{sec:conclusion} concludes the paper and makes proposals for future works.

\section{Related Works}\label{sec:related_works}

A wide range of object manipulation strategies using a swarm of robots have been studied. \cite{Tuci2018} categorise these strategies into three broad groups; pushing-only, caging, and grasping strategies. As the name suggests, the first strategy involves the miniature robots simply pushing the object with their bodies to induce its movement on a path towards the goal \citep{Sugie1995, Wang2006, Neumann2014, Chen2013, Chen2015}. The caging strategy moves the box using a similar method but distributes the robots along the entire object's circumference \citep{Brown1995, Wang_caging, Sudsang_caging}. This allows for more controlled and precise manipulations of the object’s orientation and trajectory, as its motion is controlled from all sides. Finally, the grasping strategy; rather than simply pushing against the side of an object, the robots bind themselves to the side of an object, often using an arm, grasping point or interlocking mechanism, which allows them to push and pull the object. Motivated by the way ants depend on individual-level rules to make decisions when transporting objects within a multi-agent system, the grasping strategy has been widely studied \citep{Berman2011, Wilson2014, Gelblum2016, Guo2017}. This group of manipulation strategies may also include robots that lift the object \citep{Jurt2022} or are attached to the object by means of cables, such as with aerial robots \citep{Michael2011}. We also note some works that incorporate a mix of the aforementioned strategies, such as by \cite{Ebel2021} where a dynamic algorithm distributes the robots around the side of the object in either a caging or pushing strategy dependent on the number of robots available, however, these more complex methods also often have higher computational demands to decide the optimal strategy than simply using one method.

There is also a wide range of techniques used to control and organise the swarm of miniature robots. These make use of methods ranging from finite-state machine (FSM) \citep{Kube1997, Chen2015}, decision trees \citep{Ligot2020} and, more recently, neural network-based methods that utilise evolutionary computing \citep{Alkilabi2017, Gros2009} or deep reinforcement learning \citep{Zhang2020} for synthesising the neural network robot controller. Within these control methods lies a range of implementations with varying presuppositions, robot capabilities, and levels of autonomy. \cite{Kube1997} were one of the first to propose a cooperative box manipulation task using a FSM relying purely on perceptual cues. More centralised state machines have also been proposed. A leader-follower strategy includes multiple robots pushing a box while following the trajectory of an (often human-controlled) ‘leader’ robot towards the goal \citep{Kosuge1996, Takeda2002, Wang2016, Rauniyar2021, Brown1995}. \cite{Habibi2016} designed a system where a path from the box to the goal is formulated by a number of ‘mapping’ robots that are not responsible for transporting the object. In some studies, the entire environment is visible to one camera \citep{Sugie1995} or robot \citep{Wang2006} that contributes to the centralised coordination of a global strategy over the multiple pushing robots. \cite{Neumann2014} achieved this via an external server, that sends commands to every robot.

Despite a myriad of object manipulation and robot swarm control methods, the task of object transportation using a swarm of miniature robots faces several obstacles that limit the robustness and feasibility of the above-mentioned methods in real-world applications or uncontrolled, unknown or dynamic environments. Many previous studies rely on presuppositions around some, or all, of the following \citep{Tuci2018, Jurt2022, Farivarnejad2022}: 
\begin{itemize}
    \item Predefined robot paths.
    \item Explicit robot-to-robot or controller-to-robot communication abilities.
    \item Global knowledge about an environment and/or robot position(s).
    \item A static or controlled environment.
    \item A lack of obstacles between the object and the goal.
    \item The initial positioning of the robots relative to the object or the goal.
    \item The shape, size or grasping points of the object.
\end{itemize}

Furthermore, strategies may not be robust to a fault in some element of the system. For instance, a failure of the camera or coordination server in a system reliant on global knowledge is much harder to absorb than the breakdown of a single robot in a system of twenty decentralised robots. 

While certain presuppositions and requirements seem to be an inescapable feature of many multi-robot object transportation strategies, in this study, we examine the issue of complex environments where objects are present that may block the straight-line path between the object and the goal. Existing strategies able to overcome this challenge often make use of a leader-follower approach \citep{Rauniyar2021}, wherein the leader traces a path around the obstacles. However, this comes with the assumption that the leader possesses knowledge of the global environment, or that the leader is human-controlled. Using knowledge of only the local environment has proved limiting; obstacles may occlude the vision of the simple swarm robots and hinder their ability to perceive enough of the environment to effectively execute the task at hand. While occlusion typically presents an obstacle to the object manipulation task, some works have exploited this challenge and use it as a feature of their multi-robot systems. \cite{Kube1997} and \cite{Kube2000} use a light source positioned above the goal to indicate to their simple (camera-equipped) robots the goal's position, allowing them to reposition themselves accordingly. 

More recently, \cite{Chen2013, Chen2015} proposed a novel transport strategy that utilises occlusion to determine whether a robot should push the object from its current position. Their approach is summarised as follows. The robots, the object and the goal each have distinctly different colours, and the robots, equipped with simple RGB cameras and no means to communicate with one another, all seek out and approach the object. Once at the object, they determine whether the goal is unoccluded from their position. If it is, they do not push the box from this direction (as they would push it further from the goal) and they thus opt to move around the box until the goal becomes occluded. If the goal is occluded from a robot, the robot pushes the box (towards the goal, as they are pushing the side of the object that is opposite to the goal). \cite{Chen2015} offers a mathematical proof that if the pushing robots dynamically reallocate themselves along the occluded edge of the object, the direction of the object's movement will eventually cause it to reach the goal. The success of this strategy is reliant on the object being the only item that occludes the goal from the robots, necessitating an obstacle-free path between the object and the goal. This limits the applicability of the occlusion-based strategy to environments where no, or few, obstacles are present. While noting that a single human-controlled robot with the same colour as the goal could somewhat mitigate this issue by acting as a mobile goal, \cite{Chen2015} recognise the limitation of requiring an obstacle-free path. Regardless, their occlusion-based approach demonstrates an effective method of object manipulation using simple and low-cost swarm robots without a need for global knowledge or explicit communication between the robots. 

A similar work to the occlusion-based strategy proposed by \citeauthor{Chen2015} is \citep{Alkilabi2018}. Here, the authors develop a neural network-based controller, synthesised through evolutionary computation techniques. They note that most of the best-performing strategies in their experiments exploited the occlusion of the target position by the object to determine when the robots should push the object; this strategy, developed through artificial evolution, is very similar to the explicitly-coded behaviours by \cite{Chen2015}. Furthermore, \citeauthor{Alkilabi2018} analysed the relationship between the mass and length of the objects on the task completion rate. The shorter objects are more challenging for large groups of robots due to the smaller occluded region and limited surface area upon which the robots can apply pushing force. The robots often push into each other, rather than into the object, significantly reducing the total force applied to the object. \citeauthor{Alkilabi2018} conclude that while the cylindrical shape of the robots used is a limiting factor in occlusion-based swarm transportation strategies when the surface area of the object is small, this same round shape allows the robots to move horizontally across the sides of the object.

One proposal to overcome the challenge of finding a path to the goal using only local knowledge is made by \cite{Nouyan2008}. Their method for determining a path from the object to the goal when obstacles are present entails robots forming (tightly connected) chains from the object to a position where the goal is unoccluded. Once a robot finds another robot that is part of a chain, it follows the chain until its end and either joins the chain or ends the trial if it finds an object near the (end of the) chain. \cite{Nouyan2009} extend this strategy for the object transportation task by adding behaviours that allow the robots to grasp onto the object and pull it towards the nearest member of a chain of robots connecting the prey (the object) to the nest (the goal). In both of these works, the robots indicate to each other their current state and behaviour by using different colours generated by RGB LEDs positioned around the circumference of the robots. \cite{Sugawara_pheromone} and \cite{Campo2010ArtificialPF} implemented path selection through the use of artificial pheromones emitted by each robot, acting as a simple messaging system, in response to which each robot decides whether to join, leave, or maintain a chain of robots.

\section{Methodology}\label{sec:methodology}
This study extends upon the work of \cite{Chen2013, Chen2015}, complementing their occlusion-based FSM with additional behaviours, inspired by the proposals of \cite{Nouyan2008, Nouyan2009}, that enable the robots to form stationary sub-goals along the path between the object and the goal. We construct a similar setup as by \cite{Chen2015} using simple simulated swarm robots equipped with cameras and infrared sensors and without any means of communication, pheromones, or access to any global knowledge or coordination system. Using merely visual perceptual cues, the swarm robots and their state machine are tested across multiple environments, each with different configurations of environment layouts, number of robots, and object shapes. In the remainder of this section, we will describe the robot behaviour (Section \ref{sec:robot_FSM}), simulation implementation (Section \ref{sec:sim_implementation}), robot design (Section \ref{sec:robot_description}), environments (Section \ref{sec:environments}) and shapes of the object (Section \ref{sec:shapes}) in detail. The end of this section outlines the experimental setup of this work.

The problem statement is as follows. The task consists of a bounded environment containing an object, a goal and a swarm of miniature robots that must push the object to the goal position. The environment may contain obstacles, such as walls, that block the direct path or line-of-sight between the object and the goal, but where a path between the two is still possible such that the object can fit in-between or around the walls regardless of the orientation of the object.

Following \cite{Chen2015}, we assume that the robots can distinguish the object, goal and robots in the environment by observing their colour. The dimensions of the object are such that it is large enough to completely occlude the robots' perception of the goal, should the object be in-between the robot and the goal. Unlike \citeauthor{Chen2015}, we do not assume that the goal can only be occluded to the viewpoint of any robot by the target object. We depart from this assumption in this study as our methodology aims to navigate complex environments where obstacles are present, while \citeauthor{Chen2015} presented their results in obstacle-free environments. We do, however, maintain the assumption that other robots cannot block another robot's view of the goal (and object). A further departure from the methodology by \citeauthor{Chen2015} is that we do not statically assign a (sub-)goal robot; our FSM dynamically determines which robots should become a goal, and when they should do so, while \citeauthor{Chen2015} perform an experiment where one robot is set to be a goal and it's movement is controlled by a human. We maintain no control over the individual robots; their movement and behaviours are entirely controlled by the FSM.

\subsection{Occlusion-Based State Machine} \label{sec:robot_FSM}
The behaviour of each robot is modulated by a FSM. We implement two versions of this FSM, the first replicates that of \cite{Chen2015}. This FSM serves as a baseline with which to compare our proposed FSM as we wish to ensure no performance is lost in obstacle-free settings as a result of our proposed changes. The proposed FSM augments the baseline FSM with additional behaviours that enable the robots to form sub-goal goal positions; allowing the multi-robot swarm to transport objects around obstacles present in the environment. In \autoref{fig:state_machine}, the behaviours of both FSMs are shown, where each of the states and transitions illustrated by black lines and arrows represent the behaviours of the baseline FSM, while the blue items indicate the additions made to the baseline FSM in the proposed FSM. Specifically, the state S5 represents our proposed addition to the FSM. A description of each state in \autoref{fig:state_machine}, and their respective state transition conditions, follows below:

% Figure 1
\begin{figure}[ht]
  \centering
  \includegraphics[width=\textwidth]{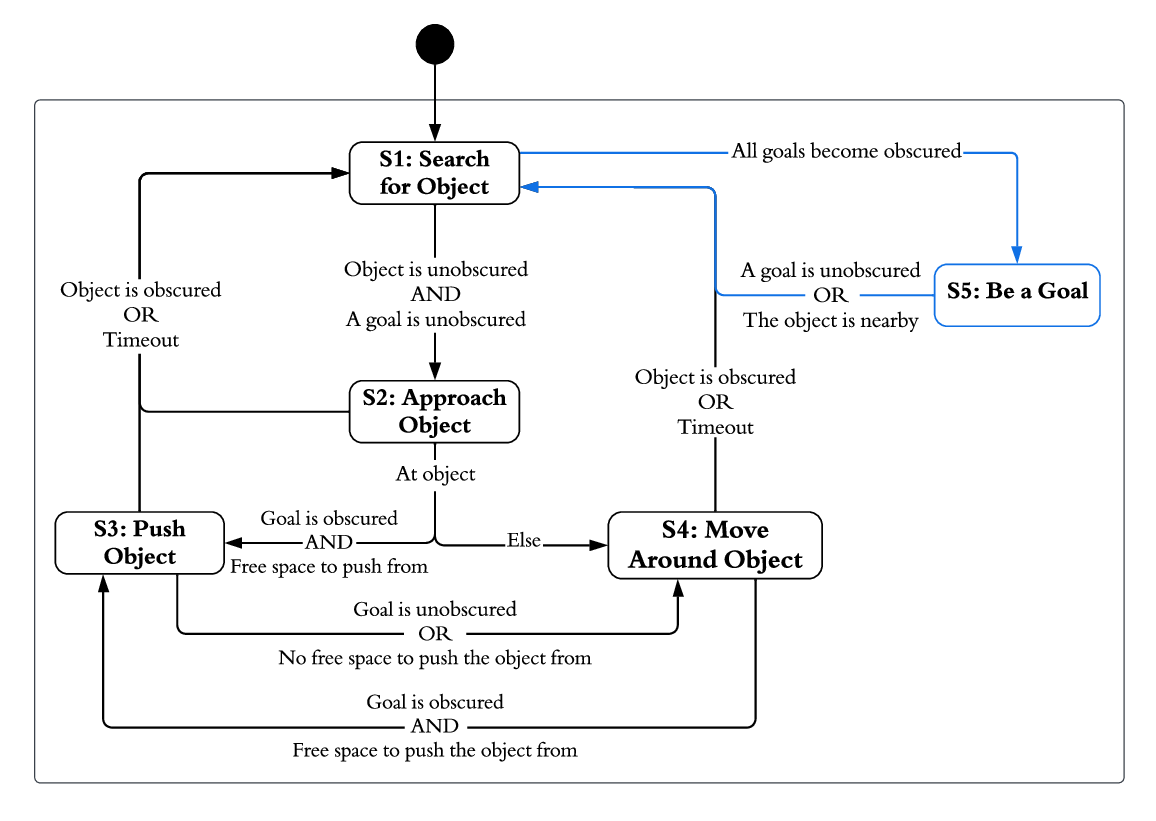}
  \caption{Finite-state machine (FSM) diagram of the proposed occlusion-based cooperative object transportation strategy. Black-coloured items indicate behaviours replicated from \citep{Chen2015}, and demonstrate the behaviour of the baseline FSM. The blue-coloured items indicate our proposed alterations. Both the blue- and black-coloured items apply to the proposed FSM.}
  \label{fig:state_machine}
\end{figure}

\begin{itemize}
    \item \textbf{S1: Search For Object}. In our implementation, the robot performs a random walk by varying the heading angle between $-\alpha$ and $\alpha$ according to a random walk process. The value of $\alpha$ is set to $0.2$ radians to introduce dynamic and less predictable movement during navigation. We opt for this approach as the random walk behaviour in \cite{Chen2015} is not described. Transitions:
    \begin{itemize}
        \item[-] $S1 \rightarrow S2$: If both the object and a (sub-)goal are unoccluded.
        \item[-] $S1 \rightarrow S5$: If all previously unoccluded (sub-)goal(s) become occluded. If the object and a (sub-)goal were simultaneously unoccluded at any point in the past, the robot will not be allowed to change to the S5 state until it has seen the goal and object with an angle greater than 90\textdegree\space between them.\footnote{This angle is calculated from the dot product of the vectors that point from the robot's position to the observed (sub-)goal, and from the robot to the object. This angle is explained in more detail in Section \ref{sec:appendixA}}
        
    \end{itemize}
    
    \item \textbf{S2: Approach Object}. The object is unoccluded and the robot moves towards it. The robot's trajectory towards the goal is controlled by the relative angle of the object to the robot, observed via the robot's cameras. Transitions:
    \begin{itemize}
        \item[-] $S2 \rightarrow S1$: If the object becomes occluded to the robot, or the robot has spent over 1 minute in this state without reaching the object.
        \item[-] $S2 \rightarrow S3$: If arrived at the object, there is no unoccluded goal and there is a free space on the object's edge to push it from.\footnote{The determination of whether or not there is a free space to push from is explained in Section \ref{sec:appendixA}.} The robot determines if it is near the object by using its RGB camera and IR sensors.
        \item[-] $S2 \rightarrow S4$: If arrived at the object, and either a goal is unoccluded or there is no free space to push from.       
    \end{itemize}
    
    \item \textbf{S3: Push Object}. The robot pushes the object perpendicular to the object's surface via the robot's point of contact with the object. Transitions:
    \begin{itemize}
        \item[-] $S3 \rightarrow S1$: If the object becomes occluded, or the robot has been in this state for over 1 minute.
        \item[-] $S3 \rightarrow S4$: If a goal is unoccluded or there is no free position from which to push the object.
    \end{itemize}
    
    \item \textbf{S4: Move Around Object}. To find a suitable position to push the object from (at which no other robots are impeding the robot's path and the goal is occluded), the robot moves around the object's perimeter. This is implemented via left- and right-wall following, with the direction of the wall following determined by the relative position of the goal and is such that the robot should not move in front of the object's path as other robots push it towards the goal. If the goal is occluded, the robot performs right-wall following around the object's perimeter. Transitions:
    \begin{itemize}
        \item[-] $S4 \rightarrow S1$: If the object becomes occluded, or the robot has been in this state for over 1 minute.
        \item[-] $S4 \rightarrow S3$: If all goals are occluded and there is a free position from which to push the object.
    \end{itemize}

    \item \textbf{S5: Be a Goal}. The robot acts as a sub-goal on the path towards the final goal. To identify itself as a goal to the other pusher robots, it turns the same colour as the goal. When leaving this state, the robot's colour returns to blue. Transitions:
    \begin{itemize}
        \item[-] $S5 \rightarrow S1$: If a goal is unoccluded, or the object is nearby.
    \end{itemize}
\end{itemize}

\begin{figure}[ht]
  \centering
  \includegraphics[width=0.8\textwidth]{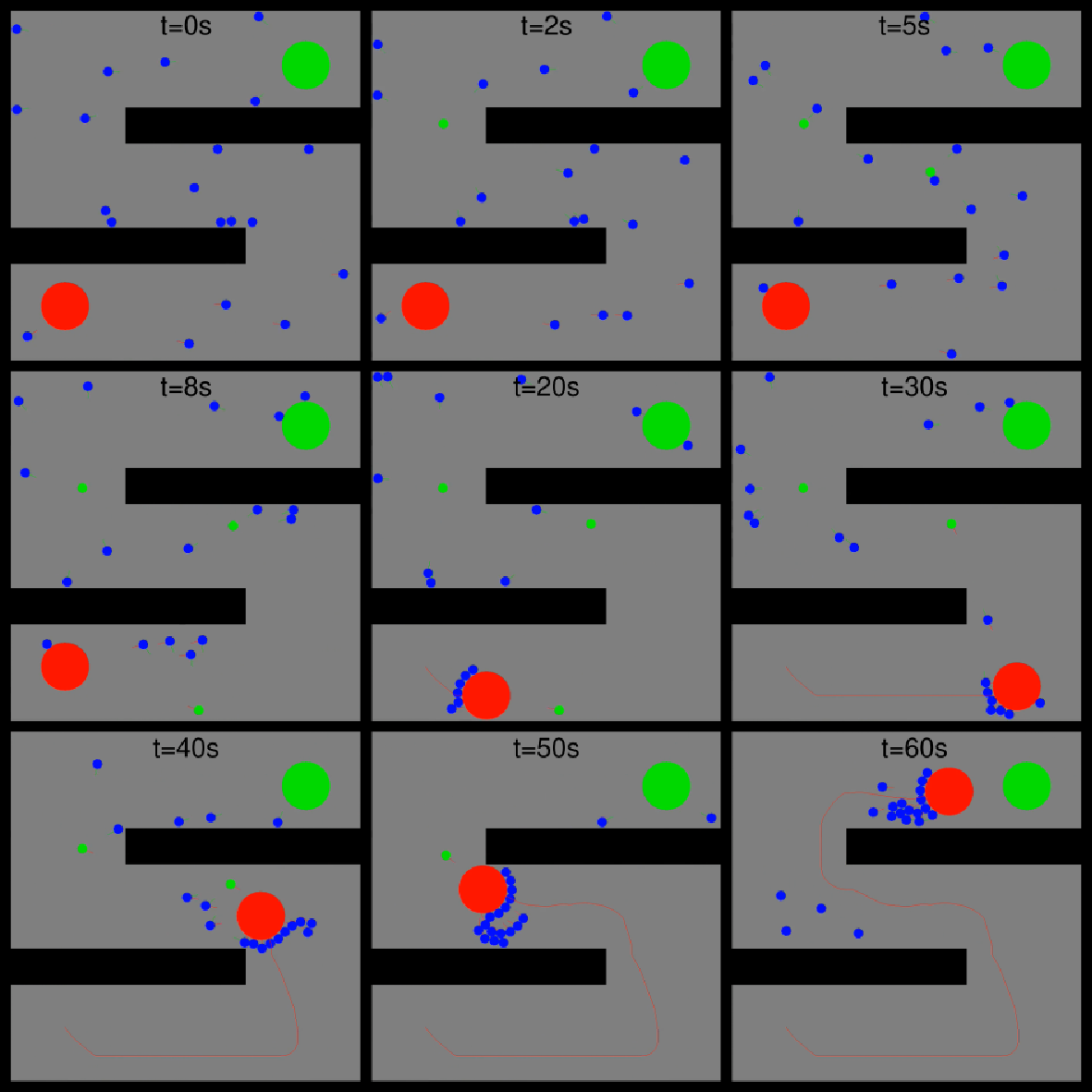}
  \caption{Time lapse of the simulation using the proposed FSM. Initially, some robots turn into sub-goals, which are later used as sub-goals by the other robots when pushing the object.}
  \label{fig:timelapse}
\end{figure}

Figure \ref{fig:timelapse} shows a time-lapse of the simulation using the proposed FSM, where 20 robots are tasked with transporting a red circular object to the green goal. The robots turn into a sub-goal when a goal/sub-goal is occluded from its cameras. Once a chain between the goal and object is created, the other robots can push the object towards the sub-goals until the goal is reached.

% Figure 2

\subsection{Simulation Implementation} \label{sec:sim_implementation}
\begin{figure}[ht]
  \centering
  \includegraphics[width=0.8\textwidth]{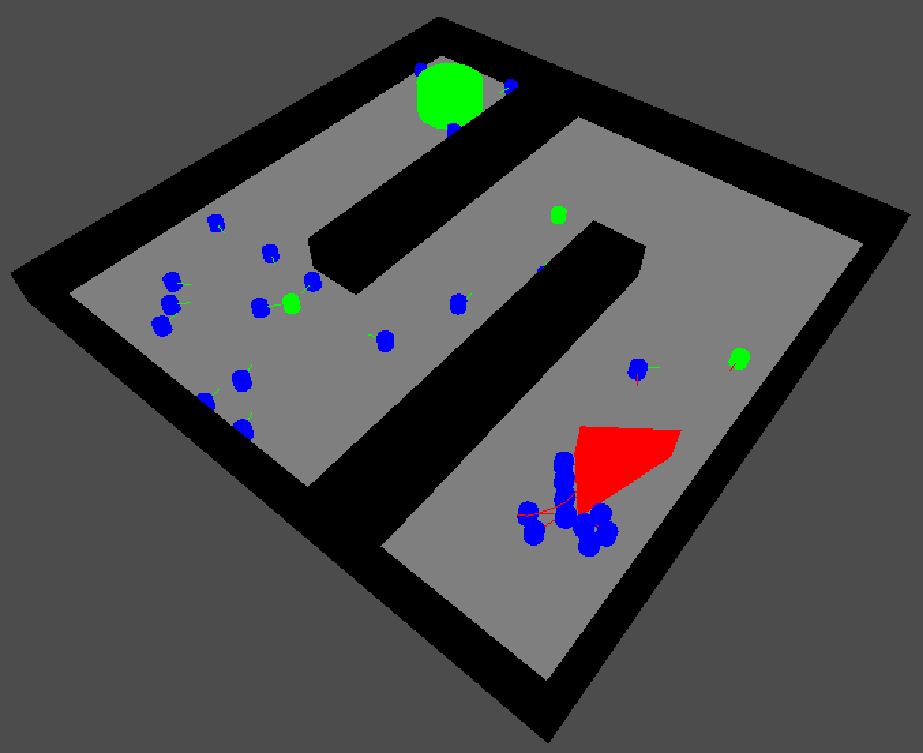}
  \caption{Simulation performed in Atta with simplified rendering. In this instance, 30 robots are moving the triangular object in the 2-Corners environment. Note how 3 of the 30 robots are forming sub-goals (green) that lead the (blue) robots to push the (red) object on a path around the walls and towards the (green) final goal.}
  \label{fig:simulation}
\end{figure}

We present the capabilities of our modifications to the occlusion-based transport strategy using a set of experiments in a simulated environment.\footnote{The source code for the simulation experiments is open-source and can be found on our research repository in \url{https://github.com/brenocq/object-transportation}.} The experiments are implemented using the Atta simulator version 0.4.0. Atta \citep{Atta2020} is an open-source simulator designed to simulate large-scale multi-robot systems. The robot's infrared sensor was implemented using ray casting in conjunction with the Box2D physics engine. The cameras were rendered using OpenGL in a 3D environment. We focused on algorithm validation, thus simplifying the rendering process by excluding shadows and textures while still obtaining meaningful sensor data, as shown in \autoref{fig:simulation}. 

% Figure 3

\subsection{Robot} \label{sec:robot_description}
\begin{figure}[ht]
  \centering
  \includegraphics[width=0.7\textwidth]{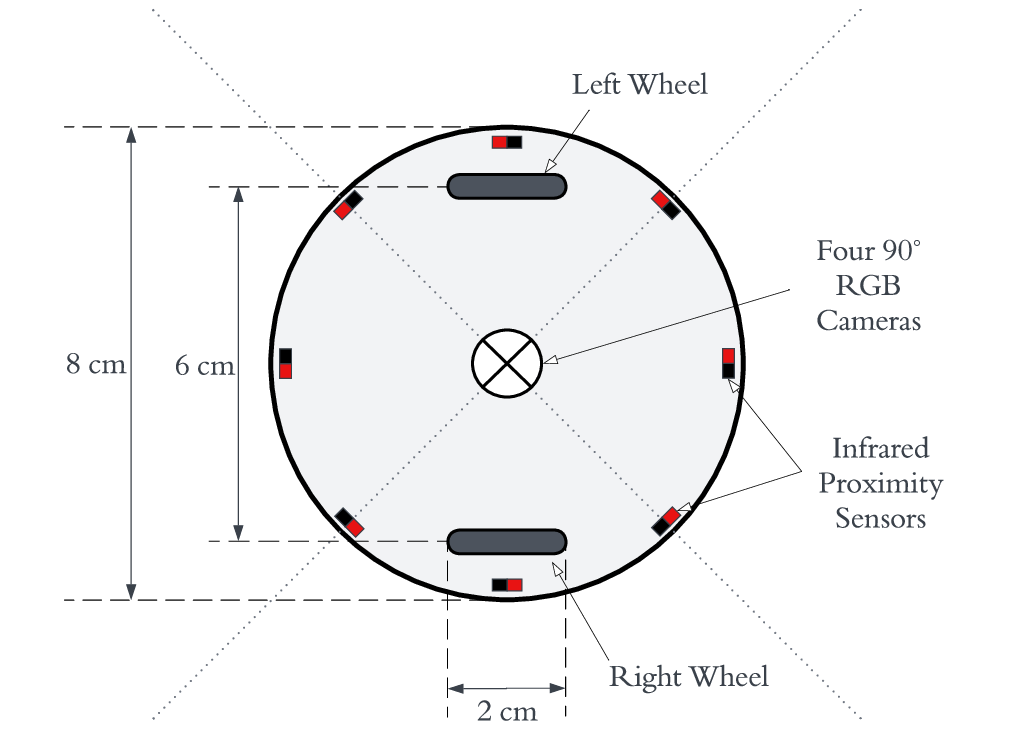}
  \caption{A top-down view schematic of the robots used in the simulation experiments indicating its dimensions and the positioning of the two differential-drive wheels, four RGB cameras, and eight infrared proximity sensors. In the simulations, the body of the robot is a bright blue, with the ability to change colour into a bright green.}
  \label{fig:robot_schematic}
\end{figure}

Within the simulated experiments, we implement a simple differential wheeled miniature robot. The robot's body consists of a blue-coloured cylinder with a diameter of 8.0 cm and a height of 6.0 cm. Each robot has a mass of 300 g. Its wheels are tucked under the bottom of the robot, are 6.0 cm apart, are both 2.0 cm in diameter, and possess a maximum speed of 0.5 m/s. The robots are not fitted with any means of explicit communication. A schematic of the robot is shown in \autoref{fig:robot_schematic}.

The robot possesses two types of sensors with which to observe its environment; four directional RGB cameras and eight infrared proximity sensors. The cameras are positioned at a point just above the centre of the top surface of the robot (9.0 cm above the ground), and each has a field of view of 90\textdegree. This configuration of the cameras enables the simulated robot to gain a 360\textdegree\space view of its surroundings, allowing the state machine to perform in optimal conditions uninhibited by sensor limitations without having to pause to rotate on the spot to observe its entire surroundings.\footnote{As would be the case with just one (forward-facing) camera.} Each camera captures an image with a resolution of $64 \times 64$ pixels, at a rate of 30 frames per second. The eight infrared proximity sensors operate at a rate of 100 measurements per second and are distributed equidistantly around the circumference of the robot, as shown in  \autoref{fig:robot_schematic}, at a height of 3.0 cm from the ground. Finally, the robots in our simulation are able to change colour. While physical robots would likely implement this ability through the use of RGB lights, for the sake of simplicity, the colour of the simulated robot's cylindrical body itself can change between a pure blue and a pure green.

% Figure 4

\subsection{Environments} \label{sec:environments}
\begin{figure}[ht]
  \centering
  \includegraphics[width=\textwidth]{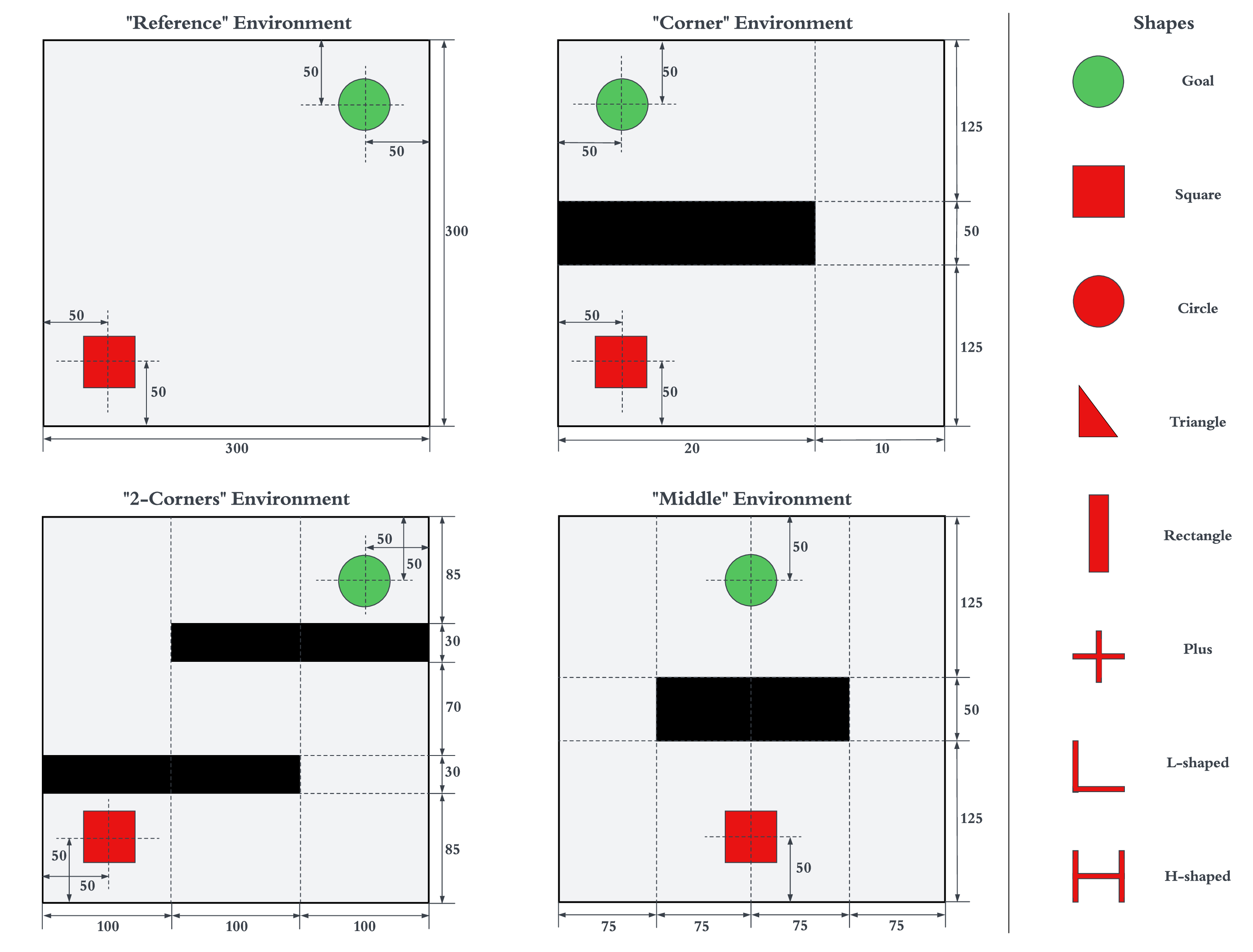}
  \caption{Maps of the four environments used in the simulated experiments. The environments are all square-shaped with bounding walls. All environments, except for the Reference environment, have walls (shown in black) that block the line-of-sight path between the goal position (indicated by the green circle) and the initial position of the object. The red square indicates this initial position, however, in our experiments, the object may instead be any of the other six object shapes described in the text. To demonstrate their proportions relative to both each other and the goal, the shapes of the objects are shown on the right side of the image. Units: cm.}
  \label{fig:maps}
\end{figure}

We create four different environments upon which to conduct our object transportation experiments. All four environments are $3 \times 3$ m in size and are bounded on each side by a 20 cm tall wall. The floor of the environment is a light grey colour, while the walls are black. Each environment contains a green, cylinder-shaped goal and a red object that are both initialised at the predefined positions illustrated in \autoref{fig:maps} at the start of every trial. 

The first environment, the Reference environment (\autoref{fig:maps}, top left), contains an environment with no obstacles. The object and the goal are placed at opposite corners of the environment, with no obstacles obstructing the line-of-sight between the object and goal. Next, the Corner environment (\autoref{fig:maps}, top right) contains one obstacle; a wall which the robots must manoeuvre the object around in order to reach the goal. There is only one possible path for the object to reach the goal, however, the goal is completely occluded from the object's starting position. This environment serves to illustrate how well our proposed FSM operates in a setting where the goal is occluded from the starting position of the object; the robots must devise a path around the wall. Similarly, the 2-Corners environment (\autoref{fig:maps}, bottom left) environment places the object and goal at opposite corners, with the additional challenge of having two walls which the robots must navigate the object around. This environment presents a more challenging version of the Corner environment, as the robots must navigate the object via a longer path around two separate walls. The primary difficulty in this environment is the length of the path that the sub-goal robots have to form to connect the path from the object to the goal. The rendering of the 2-Corners environment within a simulated experiment can also be seen in \autoref{fig:simulation}, where three (green) robots are forming a chain of sub-goals. Finally, the Middle environment (\autoref{fig:maps}, bottom right) contains just one wall that blocks the line-of-sight between the goal and the starting position of the object, however, in this environment, there are two possible paths around the sides of the obstacle. This presents a unique challenge for the proposed robot FSM, which includes no explicit decision mechanism for deciding which path (if there are multiple available) is best.

% Figure 5

\subsection{Object Shapes} \label{sec:shapes}

The objective of the robots is to move the (red) object to the position of the (green) goal. The objects have a base that can either be square, rectangular, circular, triangular, plus-, L- or H-shaped (see \autoref{fig:maps}). The physical characteristics of each of the object shapes used in our experiments are provided in \autoref{tab:shapes}. The objects are all 20 cm tall and have a mass of 5 kg, which is heavy enough such that a single robot cannot move the object alone; multiple robots (at least two) are needed for the object to move at all. The goal position is always indicated by a green circular object, with a diameter of 40 cm and a height of 20 cm.

The circular object presents the theoretically ideal case as the resulting force of the robots upon the object points directly towards the goal, however, the curved perimeter of the object increases the probability that the robots collide with one another while the object is moving. The elongated rectangle object is a challenging shape for the occlusion-based transport strategy, as the resultant force can easily cause the rectangle to rotate, rather than move directly towards the goal, depending upon which positions along the object's perimeter the robots are pushing from. The triangular object is asymmetric, which causes the resultant force vector applied to the object to rarely pass through the object's centroid. This increases the likelihood that the object rotates on the spot before achieving translational motion, which requires multiple robots pushing from the same side. The square object presents a balance of the qualities and challenges of the other four objects; it is symmetrical around its centroid and has similar dimensions to the circle object, while also having corners that would induce torque when the robots push along certain positions of the object's perimeter. The plus (or cross) shaped object poses a different challenge for the occlusion-based strategy as, unlike the other shapes, it is a concave-shaped object. This shape consists of two identical rectangle objects, with a length of 40 cm and a width of 10 cm, adjoined at their centres at a 90\textdegree\space angle to form a plus (or `cross'). This means the plus-shaped object has more sides which the robots can push from, and which may incur more directions of motion (or rotation) than the convex objects. The same is true for the L- and H-shaped objects, which are also concave-shaped, have the same maximum dimensions as the plus shape, and have the same thickness across their `arms'.

% Table 1
\begin{table}[ht]
    \centering
    \begin{tabular}{l c c c}  
        \toprule
        Shape     & Size     & Height  & Mass    \\ 
        \hline  \hline
        Square    & $40 \times 40$ cm  & 20 cm   & 5 kg    \\ %\hline
        Circle    & 40 cm diameter     & 20 cm   & 5 kg    \\ %\hline
        Rectangle    & $15 \times 60$ cm  & 20 cm   & 5 kg    \\ %\hline
        Triangle  & $30 - 40 - 50$ cm sides & 20 cm   & 5 kg    \\ 
        Plus &  $40 \times 40$ cm; $10$ cm thick `arms' & 20 cm & 5 kg \\ 
        L &  $40 \times 40$ cm; $10$ cm thick `arms'  & 20 cm & 5 kg \\ 
        H &  $40 \times 40$ cm; $10$ cm thick `arms'  & 20 cm & 5 kg \\ 
        \bottomrule
    \end{tabular}
    \caption{Characteristics of the objects used in the experiments.}
    \label{tab:shapes}
\end{table}

\subsection{Experimental Setup}

With the different environments and object shapes described above, we divide our simulated experiments into five sets. The first set of experiments takes place in the reference environment. In this environment, the baseline FSM is expected to complete the object transportation task as there are no obstacles present. This set of experiments serves to demonstrate that there is no change in performance in this setting as a result of our proposed additions of the sub-goal behaviour to the robots' FSM. The setup of this experiment is nearly identical to that of \cite{Chen2015}; there are no obstacles present within the boundaries of the environment, although there is a slight difference in the positioning of the robots at the start of a trial, which will be explained below. In the second set of experiments, we apply the proposed FSM to environments where there are obstacles present that block the line-of-sight path between the object and the goal. While current vision-based communication-free decentralised object transportation methods are unable to do so, in this set of experiments we demonstrate the potential of our proposed transportation strategy to complete the transportation task in environments with obstacles. In the third set of experiments, we demonstrate the robustness of our proposed FSM to different object shapes. This set of experiments shows that the successful performance of our proposed FSM is not conditioned on the specific shape of the object being transported, and that it is still able to complete the object transportation task if the shape is round, oblong, triangular, or concave. In the fourth set of experiments, we investigate the effect the initial positioning of the robots has on the task completion rate. More specifically, we compare three different robot position initialisation methods that are used at the start of each trial: all robots are placed near the goal, all robots are placed near the object, and all robots are placed randomly anywhere in the environment.

In the final experiment, we benchmark our proposed method against a strategy that uses a simulated teleoperated goal robot. In this benchmark strategy, one robot is assigned to act as a green sub-goal, and is driven along a predefined path between the object's starting position and the goal while maintaining a constant distance of 0.5 m away from the object. The path is calculated using Dijkstra's algorithm, with the condition that the path must always be at least the width and depth of the object (whichever is larger) away from the environment's walls. All other robots are controlled using the baseline FSM (and thus can not form additional subgoals). The assumptions made in this STR strategy mark a significant departure from those used in our proposed method; the teleoperation assumes that the environment's layout is known beforehand, while our method does not require this knowledge. Thus, the simulated teleoperation strategy serves as a heuristic for the `ideal' problem scenario. This method is inspired by a very similar experiment carried out by \citeauthor{Chen2015}, wherein they use the same method except for that the statically-assigned goal robot in their work is controlled by a human, while in this work the robot is self-controlled. We opt to use this method as it ensures that the distance between the mobile sub-goal robot and the object remains constant, and such that the path the robot takes towards the final goal is as efficient as possible at each timestep throughout the experiment.

\begin{table}[ht]
    \centering
    %\small
    \begin{tabular}{m{0.18\textwidth}  m{0.13\textwidth} m{0.09\textwidth} m{0.13\textwidth} m{0.11\textwidth}  m{0.18\textwidth}}
    %\toprule
        \toprule
        \textbf{Experiment} & \textbf{Environment} & \textbf{Object Shapes} & \textbf{Robot FSM} & \textbf{Number of Robots} & \textbf{Robot Position Initialisation} \\
        \hline
        \hline
        \textbf{Obstacle-Free Environment} & Reference & Square & Baseline, Proposed & 5, 10, 15, 20, 30 & Random \\
        \hline
        \textbf{Environments with Obstacles} & Corner, 2-Corner, Middle & Square & Proposed & 5, 10, 15, 20, 30 & Random  \\
        \hline
        \textbf{Different Shapes} & All & All & Proposed & 20 & Random  \\
        \hline
        \textbf{Initialisation Strategy} & All & Square & Proposed & 20 & Near Goal, Near Object, Random  \\
        \hline
        \textbf{Teleoperation Benchmark} & All & Square & Proposed, Teleoperated & 20 & Random  \\
        \bottomrule
    %\bottomrule
    \end{tabular}
    \caption{Summary of the configuration of each set of experiments. Rows: the experiments. Columns: experiment parameters.}
    \label{tab:experiments}
\end{table}

Within each set of experiments, we utilise various configurations of the task. As described above, we may utilise different environments, object shapes and FSMs to control the robots, but we observe the impact of having different numbers of robots present in the environment on the robots' task performance. The possible configurations of these parameters for each of the five sets of experiments are summarised in \autoref{tab:experiments}. For each possible combination of the four parameters shown in this table (that is, one value per column), we run 50 trials. Thus, for the experiments in an obstacle-free environment, we use 500 trials as we have two possible FSMs and five possible values for the number of robots present in the environment. Using the same reasoning, we use 750 trials for the experiments using environments with obstacles (three possible environments, five numbers of robots), 1400 for the experiments using different shapes (four environments, seven different object shapes), 600 for the initialisation strategy experiments (four environments, three initialisation methods), and 400 trials for the teleoperation comparison (four environments, two robot FSMs).\footnote{Note that we do not re-run a set of 50 trials if a certain combination of configurations is shared between the sets of experiments. For example, the same results for the trials of the Corner environment with 20 proposed FSM robots, the square object and random initialisation are used in all but the first set of experiments, as that configuration is used within each of those sets.}

Unless otherwise specified, the robots' initial positions on each trial are generated randomly by a uniform distribution over the surface of the environment, but such that their positions do not overlap with the obstacle, goal, walls or another robot.\footnote{This is a slight deviation from the methodology of \cite{Chen2015}, as they initialise their robots within a smaller region in-between the object and the goal, while our initialisation spreads the robots out over the entire environment.} The object is also initialised with a randomised orientation on the starting position indicated by the red square in \autoref{fig:maps}. Each trial ends once one of the following conditions are met:
\begin{enumerate}
    \item When the distance between the centroids of the object and the goal falls below a distance threshold, the trial is considered successful.
    \item A time limit of 20 minutes is reached, and the trial is considered unsuccessful.
\end{enumerate}

The distance thresholds are dependent upon the shape used in that specific trial. We define the threshold as the maximum possible distance between the centroids of the two shapes while they are still colliding, plus a small margin of 5 cm. The precise thresholds for each of the object shapes are described below:
\begin{itemize}
    \item[-] \textit{Square}: The sum of half of the diagonal length of the square object and the radius of the goal, plus 5 cm.
    \item[-] \textit{Rectangle}: The sum of half of the rectangle's diagonal length and the radius of the goal, plus 5 cm.
    \item[-] \textit{Circle}: The sum of the object's radius and the goal's radius, plus 5 cm.
    \item[-] \textit{Triangle}: The sum of half of the triangle's longest side length and the goal's radius, plus 5 cm.
    \item[-] \textit{Plus, L} and \textit{H}: The same ending condition as the square object, as these shapes have the same maximum length and width as the square object.
\end{itemize}

Finally, when analysing the results of the trials, three metrics are used. First, the completion rate indicates the percentage of the 50 trials using that configuration of parameters wherein the object transportation task was completed. Second, the task completion duration measures the amount of time, in seconds, taken for the robots to complete the task. Finally, we record the path efficiency of successful trials. The path efficiency for each trial is defined as
\begin{equation}
    PE = \frac{d_{\mathrm{min}}}{d},
\end{equation}
where $d$ is the length of the path travelled by the object's centroid throughout one trial, and $d_{\mathrm{min}}$ is the length of the shortest possible path of the centroid between the object's initial position and the radius around the goal defined by the distance thresholds above. $d_{\mathrm{min}}$ differs per environment layout and is also dependent on the object's shape.\footnote{The waypoints of the shortest possible path account for the object being rotated against the corners of the walls, as this path defines the shortest path that the centroid has to travel to reach the goal. The optimal path differs per shape as the centroids of narrower objects can trace closer to the corners of the walls in the environments, and because the shapes have different distance thresholds for the task completion criteria.} Both $d$ and $d_{\mathrm{min}}$ are defined purely by the translational motion of the object's centroid, and do not incorporate the amount of rotation experienced by the object throughout the trial. The optimal path is therefore also independent of the number, or positions, of waypoints indicated by the robot sub-goals. The paths of the object in each trial across all five experiments in this paper are presented in the supplementary material. For the latter two metrics, we only present the results of the successful trials. In the following five sections, we describe the results of each set of experiments outlined in \autoref{tab:experiments}.

%%%%%%%%%%%%%%%%%%%%%%%%%%%%%%%%%%%%%%%%%%%%%%%%%%%%%%%%%%%%%%%%%%%%%%%%%%%%%%
%%%%%%%%%%%%%%%%%%%%%%  EXPERIMENTS %%%%%%%%%%%%%%%%%%%%%%
%%%%%%%%%%%%%%%%%%%%%%%%%%%%%%%%%%%%%%%%%%%%%%%%%%%%%%%%%%%%%%%%%%%%%%%%%%%%%%

%%%%%%%%%%%%%%%%%%%%%%%
%%    EXPERIMENT 1   %%
%%%%%%%%%%%%%%%%%%%%%%%

\section{Experiments in the Obstacle-Free Environment} \label{sec:Obstacle-Free_map_Experiments}

\begin{figure}[ht]
  \centering
  \includegraphics[width=\textwidth]{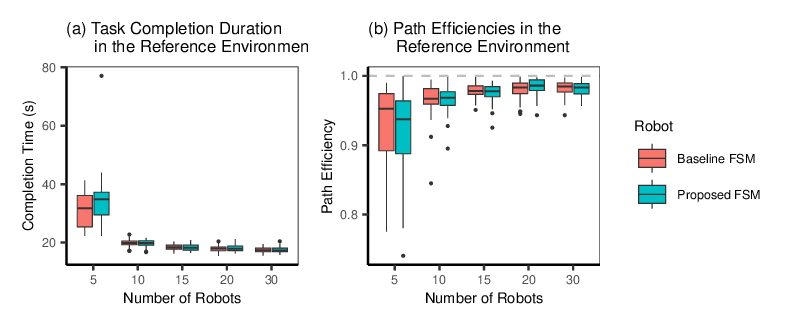}
  \caption{Task completion duration (\textbf{a}) and path efficiency (\textbf{b}) results of the simulated trials in the obstacle-free Reference environment.}
  \label{fig:exp1_plots}
\end{figure}

The first experiments concern the baseline and proposed FSMs when applied to the Reference environment, where no obstacles are present. For both FSMS, the task completion rate is 100\%, regardless of the number of robots used. %The results of the two FSMs in this environment are shown in \autoref{tab:exp_1}. 
The task completion times and path efficiencies for each configuration of the number of robots are also shown in \autoref{fig:exp1_plots}. We observe that the time taken to complete the task is consistent between the baseline and the proposed FSM. Furthermore, the path efficiency of the object is not affected by which FSM is used. From the results of this set of experiments, we conclude that our proposed changes to the baseline FSM do not result in a loss of performance in environments the baseline FSM was designed for: environments where no obstacles are present that occlude the line-of-sight between the object's position and the goal.

% Figure 6

%%%%%%%%%%%%%%%%%%%%%%%
%%    EXPERIMENT 2   %%
%%%%%%%%%%%%%%%%%%%%%%%

\section{Experiments in Environments with Obstacles} \label{sec:Obstacle_map_Experiments}

In this set of experiments, we employ our proposed FSM in environments where obstacles block the direct path between the goal and the object. This type of environment is currently out of reach of current methods in vision-based communication-free decentralised object transportation methods, which are unable to navigate the object around obstacles that impede the line-of-sight between the goal and the object.

\begin{table}[ht]
    \centering
    \begin{small}
    \begin{tabular}{l c c d{3.2} d{3.2} c d{1.2} d{1.2}} 

        \toprule
        \multirow{2}{*}{\centering Environment}     & \multirow{2}{*}{\centering \shortstack{Number of \\ Robots}}       & \multirow{2}{*}{\centering \shortstack{Completion \\ Rate}}    & \multicolumn{2}{c}{Completion Time (s)} & & \multicolumn{2}{c}{Path Efficiency}    \\ \cmidrule{4-5} \cmidrule{7-8}
        &  &  & \multicolumn{1}{c}{Mean} & \multicolumn{1}{c}{SD} & & \multicolumn{1}{c}{Mean} & \multicolumn{1}{c}{SD}
        \\
        
        \midrule
        \midrule
        Corner & 5  &   4\% & 617.88 &  87.82 & & 0.80 & 0.02 \\
               & 10 &  94\% & 283.10 & 295.47 & & 0.78 & 0.08 \\
               & 15 & 100\% & 103.14 & 102.26 & & 0.79 & 0.09 \\
               & 20 & 100\% &  65.11 &  35.18 & & 0.82 & 0.06\\
               & 30 & 100\% &  52.09 &  16.50 & & 0.82 & 0.09 \\
        \midrule
        2-Corners & 5 & 0\% & - & - & & - & - \\
                  & 10 & 46\% & 564.30 & 285.67 & & 0.70 & 0.09\\
                  & 15 & 88\% & 367.70 & 313.76 & & 0.69 & 0.12 \\
                  & 20 & 94\% & 160.59 & 171.86 & & 0.72 & 0.08 \\
                  & 30 & 98\% & 108.76 & 146.96 & & 0.73 & 0.12 \\
        \midrule
        Middle & 5  &  6\% & 705.99 & 228.86 & & 0.70 & 0.03 \\
               & 10 & 52\% & 270.33 & 309.31 & & 0.69 & 0.14 \\
               & 15 & 64\% & 182.79 & 167.90 & & 0.68 & 0.15 \\
               & 20 & 76\% & 227.61 & 292.42 & & 0.67 & 0.16 \\
               & 30 & 76\% & 146.79 & 168.48 & & 0.67 & 0.17 \\

        \bottomrule
    \end{tabular}
    \end{small}
    \caption{Results of the proposed robot FSM with the square object in the three environments which have obstacles in-between the object's starting position and the goal.}
    \label{tab:exp_2}
\end{table}

The results of the proposed FSM in these challenging environments are shown in \autoref{tab:exp_2}. Overall, we observe from the simulated experiments that when there are just 5 or 10 robots present in the environment there is often not enough force applied to the object to move it, as there may be just one or two robots pushing the object, while the rest of the robots are either walking around the environment trying to find the object or are acting as a sub-goal themselves. The task is only consistently completed whenever there are enough robots in the environment left over to push the object, once a chain of sub-goals (between the object and the goal) has been formed. This is especially necessary for the 2-Corners environment, where multiple robots must convert into the sub-goal state (S5) for the object to be able to trace its way around the two walls. Here, the probability of all of the robots finding the object through random walking is also reduced, as there is no explicit path planning or navigation strategy with which the robots can self-organise to find the object more efficiently than the current random walk implementation. For the Corner environment, at least 15 robots are required to consistently complete the task, while 20 robots were needed for the 2-Corners and Middle environments.

Still, the completion rate on these three environments is not 100\%. In the 2-Corners environment (with at least 20 robots), the task is not completed in one or three of the fifty trials. We attribute this error to the object getting stuck in the corner of the environment; a scenario that the robots do not have an explicit mechanism to escape from. We verify that this occurrence is the reason behind the failed trials by observing the paths that the objects took (Figures 2 and 3 in the supplementary material), where we notice the square object becomes stuck in the bottom left corner of the environment once and also the bottom right corner once. The Middle environment remains the most challenging; the completion rate never surpasses 76\%. When watching the trials, we notice that the object can occasionally become stuck between sub-goal robots on opposite sides of the object. As there is no communication between the robots, having multiple sub-goals in opposite directions to the object (around either side of the wall in the centre of the Middle environment) presents a limitation of task completion when there are multiple possible paths for the object to take to reach the goal. Our current work contains no explicit algorithm or strategy for handling the possibility of multiple (symmetrical) paths, thus, on the Middle environment, some time is lost when the robots have to choose between two paths of equal length.

% Figure 7
\begin{figure}[ht]
  \centering
  \includegraphics[width=\linewidth]{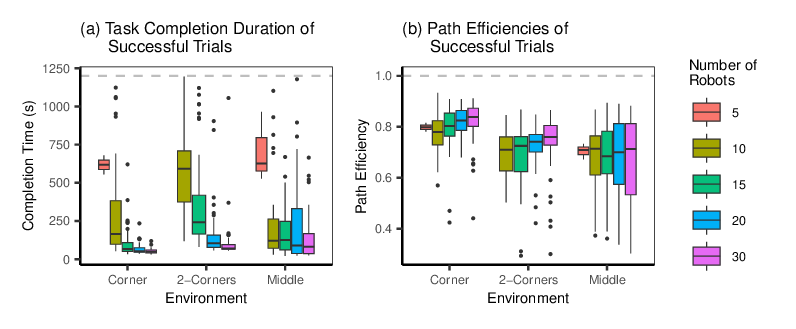}
  \caption{Box plots describing  (\textbf{a}) the task completion durations and (\textbf{b}) the object path efficiencies of the experiments with the proposed FSM in environments where obstacles are present.}
  \label{fig:exp2_plots}
\end{figure}

The task completion duration and path efficiencies for this set of experiments are shown in \autoref{fig:exp2_plots}. Again, only the results of trials where the object transportation task is successfully completed are shown in these figures. We observe that across all environments, the task completion duration generally declines as the number of robots increases. This is due to more robots being available to apply force to the object; moving the object around the environment faster. With over 20 robots, the Corner environment is completed in the shortest time, while the 2-Corners and Middle environments take the longest to complete. It is interesting to note that while the optimal path in the Middle environment is roughly half that of the 2-Corners environment, the Middle environment takes a similar amount of time to complete (for $\geq15$ robots). During the trials, we observed that the object would often be slid back and forth between the two possible paths around either side of the wall, due to sub-goals being present on either side of the wall simultaneously. This is also reflected in the Middle environment consistently having the lowest mean path efficiency.

The path efficiency per environment does not appear to be largely affected by the number of robots. We find that while the objects are pushed along their paths faster if there are more robots present in the environment, the shape of the path taken by the object is not contingent on the number of robots as the mean and variance of the path efficiencies (of successful trials) remain stable across different numbers of robots. Again, the paths of the objects throughout the trials are shown in the supplementary material.

%%%%%%%%%%%%%%%%%%%%%%%
%%    EXPERIMENT 3   %%
%%%%%%%%%%%%%%%%%%%%%%%

\section{Experiments with Different Shapes} \label{sec:Different_Shape_Experiments}

To further assess the robustness of our proposed implementation of the multi-robot occlusion-based object transportation strategy, we examine the results of our experiments on all four different environments (environments with and without obstacles) using different object shapes. The results of these experiments are shown in \autoref{tab:exp_3}.

\begin{table}[ht]
    \centering
    \begin{small}
    \begin{tabular}{l l c d{3.2} d{3.2} c d{1.2} d{1.2}}  
        \toprule
        \multirow{2}{*}{\centering Environment}     & \multirow{2}{*}{\centering Shape}       & \multirow{2}{*}{\centering \shortstack{Completion \\ Rate}}   & \multicolumn{2}{c}{Completion Time (s)} & & \multicolumn{2}{c}{Path Efficiency}    \\ \cmidrule{4-5} \cmidrule{7-8}
        &  &  & \multicolumn{1}{c}{Mean} & \multicolumn{1}{c}{SD} & & \multicolumn{1}{c}{Mean} & \multicolumn{1}{c}{SD}
        \\
        \midrule
        \midrule
        Reference & Circle    &  100\% & 20.97 &   0.88 & & 0.99 & 0.01 \\
                  & Square    &  100\% & 18.06 &   1.15 & & 0.99 & 0.01 \\
                  & Rectangle &  100\% & 26.02 &   5.26 & & 0.88 & 0.06 \\
                  & Triangle  &  100\% & 25.54 &   6.92 & & 0.91 & 0.07 \\
                  & Plus      &  100\% & 35.43 &  16.55 & & 0.93 & 0.02 \\
                  & L         &   96\% & 64.03 & 114.02 & & 0.91 & 0.09 \\
                  & H         &  100\% & 51.41 &  35.02 & & 0.93 & 0.09 \\
        \midrule
        Corner    & Circle    &  100\%  &  53.49 &  14.26 & & 0.90 & 0.05 \\
                  & Square    &  100\%  &  65.11 &  35.18 & & 0.82 & 0.06 \\
                  & Rectangle &  100\%  & 197.42 & 165.02 & & 0.70 & 0.09 \\
                  & Triangle  &   98\%  & 108.94 &  67.38 & & 0.72 & 0.09 \\
                  & Plus      &  100\%  &  94.69 &  40.18 & & 0.83 & 0.07 \\
                  & L         &   94\%  & 132.91 &  92.07 & & 0.76 & 0.13 \\
                  & H         &   80\%  & 402.97 & 191.13 & & 0.64 & 0.11 \\
        \midrule
        2-Corners & Circle    &  100\% &  97.72 &  46.97 & & 0.78 & 0.11 \\
                  & Square    &  94\%  & 160.59 & 171.86 & & 0.72 & 0.08 \\
                  & Rectangle &  58\%  & 563.69 & 335.22 & & 0.61 & 0.09 \\
                  & Triangle  &  88\%  & 312.46 & 228.89 & & 0.55 & 0.14 \\
                  & Plus      &  96\%  & 169.33 & 181.92 & & 0.70 & 0.16 \\
                  & L         &  82\%  & 285.70 & 224.47 & & 0.70 & 0.13 \\
                  & H         &  46\%  & 760.21 & 208.57 & & 0.60 & 0.10 \\
        \midrule
        Middle    & Circle    &  90\%  & 149.83 & 196.32 & & 0.77 & 0.12 \\
                  & Square    &  76\%  & 227.61 & 292.42 & & 0.67 & 0.16 \\
                  & Rectangle &  58\%  & 334.67 & 328.75 & & 0.48 & 0.15 \\
                  & Triangle  &  96\%  & 165.84 & 197.57 & & 0.55 & 0.16 \\
                  & Plus      &  76\%  & 169.33 & 181.92 & & 0.70 & 0.16 \\
                  & L         &  90\%  & 204.30 & 209.28 & & 0.62 & 0.20 \\
                  & H         &  80\%  & 339.00 & 263.95 & & 0.58 & 0.19 \\
                  
        \bottomrule
    \end{tabular}
    \end{small}
    \caption{Results of the proposed robot FSM in all four environments, using differently-shaped objects.}
    \label{tab:exp_3}
\end{table}

% Figure 8
\begin{figure}[ht]
    \centering
    \includegraphics{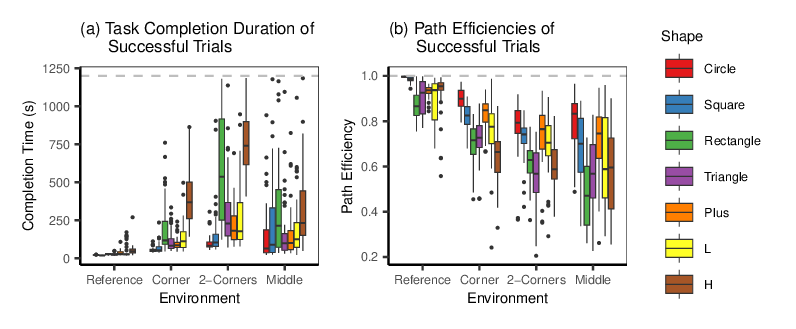}
    \caption{Results of the experiments using differently-shaped objects on all of our environments. (\textbf{a}) shows the durations of successful trials, while (\textbf{b}) presents the path efficiencies of those same trials. All trials use 20 robots controlled by the proposed FSM.}
    \label{fig:exp3_plots}
\end{figure}

The proposed FSM is able to achieve a completion rate of over 75\% in every environment with most of the shapes. Out of the 28 combinations of object shape and environment, only three combinations have a completion rate lower than this. In the Middle and 2-Corners environments, the rectangle object has a completion rate of just 58\%, while the H-shaped object had a completion rate below 50\% in the 2-Corners environment. In \autoref{fig:exp3_plots}, we observe that the rectangle and H-shaped objects demonstrate the largest variances in task completion duration. In environments with obstacles, these two shapes also accounted for the highest mean completion time and one of the lowest mean path efficiencies. While observing the trials, we notice that the swarm of robots can occasionally get stuck and struggle to progress these two objects further towards the goal whenever the long edge of the rectangle or H-shape is flush against one of the walls (with no space for any robots between the object and the wall).\footnote{The long edge of the object pressing against one of the two interior walls of the 2-Corners environment is a much more common occurrence than it being alongside one of the walls that outline the four sides of the environment.} While this also occasionally occurs with the square and L-shape objects, this occurrence is a larger hindrance to the rectangle object, as its long shape makes it more difficult to rotate the object back off of the wall by pushing against one of the corners on the opposite side of the shape. The concave regions inside of the H-shape are also more likely to get caught on the edges of the walls in the Corner and 2-Corners map. The robots are not equipped with any explicit strategy for moving the object away from a wall, nor preventing the object’s sides from being directly adjacent to a wall (without any space in between the wall and the object).

The challenge of transporting concave objects (the plus, L and H shapes) is relatively well dealt with. Despite the exception of the H-shape on the 2-Corners map described above, the success rate of all other trials with the concave objects achieves a success rate of at least 76\%. The H-shape had the highest mean completion times on the three environments with obstacles, and at least the second-lowest path efficiency on every map. This suggests that this object shape was the most challenging for our proposed FSM to transport. We attribute this to the occasions of the H-shape getting stuck against a wall discussed above, as well as excessive rotations of the object when the robots are pushing it. Due to its concave shape, force applied to the H-shape from inside the two concave regions generally applies more torque to the object than translational force. This causes the path efficiency to decrease, as the object rotates off of a direct line to the next (sub-)goal, and the slower translational movement also causes the task completion duration to increase. We also note that the completion rate of the L-shaped object falls below 100\% in the Reference environment. When observing the experiments, we find that the 2 failed trials of this combination are caused by a large amount of robots being caught inside of the concave region inside the L-shape, which greatly inhibits the movement of the object, and causes it to get stuck against a wall. For the plus and L-shapes on the obstacle environments, we do not observe a large discrepancy in the completion rates, completion times or path efficiencies between these two concave objects and the four convex objects. Thus, we find that our proposed FSM was able to transport both concave and convex objects successfully.

However, in the Reference environment, the mean completion times of the experiments with concave objects are higher than the convex objects, while the variance is considerably greater. This is despite the path efficiencies between the convex and concave objects being comparable. When reviewing the trial footage, we notice that occasionally some (2 or 3) robots may get stuck against the inner faces of the concave objects while the other robots are pushing the object towards the goal. This suggests a limitation of the wall-following algorithm used to reposition the robots to the side of the object opposite from the goal. As all 20 robots can easily find and approach the object in the reference environment; the (concave) objects can be moved towards a goal at a faster rate than the robots can use their wall-following algorithm to move away from the side of the object facing the goal (thus, they are somewhat trapped against the front of the object, inhibiting its speed). This does not greatly affect the path efficiency, as the object can still move close to the ideal path, but does have a large impact on the completion time. 

Out of the concave objects, the triangle shape presented the second-largest challenge, after the rectangle object. The triangle resulted in high completion rates, but relatively low path efficiencies, with high variances. We attribute this variance to the asymmetry of the triangle object, which causes more rotation of the object when the robots are pushing it than occurs with the other concave object shapes, making the triangular object more difficult and unpredictable to move. The triangle, rectangle, and H-shaped objects consistently have the lowest mean path efficiency. As expected, the circle and square objects are easier to transport; having lower task completion times, higher path efficiencies, and less variance in both metrics. All shapes demonstrate their lowest path efficiencies in the Middle environment, likely due to the availability of multiple possible paths, as outlined in the previous experiment.

\section{Experiments with Different Robot Starting Positions} \label{sec:Starting_Position_Experiments}

% Figure 9
\begin{figure}[ht]
    \centering
    \includegraphics{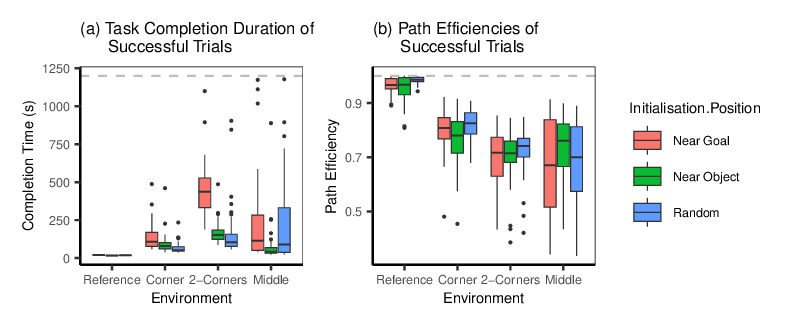}
    \caption{Results of the experiments using different initialisation strategies for positioning the robots at the start of each trial. All trials use 20 robots controlled by the proposed FSM.}
    \label{fig:exp4_plots}
\end{figure}

\begin{table}[ht]
    \centering
    \begin{small}
    \begin{tabular}{l l c d{3.2} d{3.2} c d{1.2} d{1.2}} 
        \toprule
        \multirow{2}{*}{\centering Environment}     & \multirow{2}{*}{\centering \shortstack{Initial \\ Position}}       & \multirow{2}{*}{\centering \shortstack{Completion \\ Rate}}  & \multicolumn{2}{c}{Completion Time (s)} & & \multicolumn{2}{c}{Path Efficiency}    \\ \cmidrule{4-5} \cmidrule{7-8}
        &  &  & \multicolumn{1}{c}{Mean} & \multicolumn{1}{c}{SD} & & \multicolumn{1}{c}{Mean} & \multicolumn{1}{c}{SD}
        \\
        \midrule
        \midrule
        Reference & Near Goal   &  100\% & 20.90 & 0.88 & & 0.97 & 0.03 \\
                  & Near Object &  100\% & 16.66 & 0.86 & & 0.95 & 0.05 \\
                  & Random      &  100\% & 18.06 & 1.15 & & 0.99 & 0.01 \\
        \midrule
        Corner    & Near Goal   &  100\% & 136.70 & 85.76 & & 0.80 & 0.08 \\
                  & Near Object &  100\% &  90.55 & 63.65 & & 0.77 & 0.09 \\
                  & Random      &  100\% &  65.11 & 35.18 & & 0.82 & 0.06 \\
        \midrule
        2-Corners & Near Goal   &  96\% & 445.01 & 169.63 & & 0.69 & 0.10 \\
                  & Near Object &  92\% & 164.73 &  70.05 & & 0.70 & 0.09 \\
                  & Random      &  94\% & 160.59 & 171.86 & & 0.72 & 0.08 \\
        \midrule
        Middle    & Near Goal   &  78\% & 241.65 & 297.59 & & 0.66 & 0.18 \\
                  & Near Object &  90\% &  79.50 & 133.60 & & 0.74 & 0.11 \\
                  & Random      &  76\% & 227.61 & 292.42 & & 0.67 & 0.16 \\
        \bottomrule
    \end{tabular}
    \end{small}
    \caption{Results of the experiments using different robot position initialisation approaches on all four environments (using 20 robots and the square shape in all trials).}
    \label{tab:exp_4}
\end{table}

In this set of experiments, we discuss the influence of the positioning of the robots at the initialisation of each trial. We compare three methods; initialising the robots near the goal, initialising the robots near the object, and distributing the robots randomly across the entire environment (only the random strategy was used in all other experiments). The completion rate, completion time and path efficiencies are given in \autoref{tab:exp_4}, while the latter two metrics are also displayed in \autoref{fig:exp4_plots}. The results of these strategies show some nuances in the different environments. In the Reference environment, the completion time is slightly improved by placing the robots near the object, which reduces the time needed for most of the robots to approach the object before they can begin pushing it. In the Corner environment, the random positioning method yielded the lowest completion time and highest path efficiency. The spread of the robots across the entire environment improved the speed at which a chain of sub-goals was established, as the waiting time for one robot to find a path between the object and goal did not require the robot to walk the distance between the goal or object and the end of the wall present in the Corner environment.

In the 2-Corners environment, initialising the robots randomly or near the object yielded similar results, but placing them near the goal led to a slightly lower completion rate and a much higher completion time. While we believe both of the non-random initialisation methods are equally challenging in establishing a chain of subgoals, placing the robots near the goal reduced the number of robots near the object; leading to fewer robots being able to push the robot at the beginning of the trial (causing the object to move slower) as well as it taking longer for more robots to find and approach the object before they can supply more pushing force. Finally, in the Middle environment, placing the robots near the object yielded the highest completion rate, the lowest completion time and the highest path efficiency. We attribute this to similar reasons as to why this method was also the best-performing on the Reference environment; there are more robots accumulated near the object prepared to push it than in the other two strategies. This, paired with the lower risk that two simultaneous paths are formed (which is higher with the random initialisation method, which may place multiple robots near to where a sub-goal can or should be formed).

\section{Comparison to a Simulated Teleoperated Strategy} \label{sec:Teleop_Experiment}

\begin{table}[ht]
    \centering
    \begin{small}
    \begin{tabular}{l c c c c c c c}  
        \toprule

        \multirow{2}{*}{\centering Environment}     & \multirow{2}{*}{\centering Controller}       & \multirow{2}{*}{\centering \shortstack{Completion \\ Rate}}    & \multicolumn{2}{c}{Completion Time (s)} & & \multicolumn{2}{c}{Path Efficiency}    \\ \cmidrule{4-5} \cmidrule{7-8}
        &     &   & Mean & SD & & Mean & SD 
        \\
        \midrule
        \midrule
        Reference & Proposed     & 100\%  &  18.06 & 1.15 & & 0.99 & 0.01 \\
                  & Teleoperated & 100\%  &  17.86 & 0.84 & & 0.99 & 0.01 \\
        \midrule
        Corner & Proposed        & 100\%  &  65.11 & 35.18 & & 0.82 & 0.06 \\
               & Teleoperated    & 100\%  &  42.65 & 32.33 & & 0.87 & 0.04 \\
        \midrule
        2-Corners & Proposed     &  94\%  & 160.59 & 171.86 & & 0.72 & 0.08 \\
                  & Teleoperated & 100\%  & 100.11 &  67.93 & & 0.81 & 0.02 \\
        \midrule
        Middle & Proposed        &  76\%  & 227.61 & 292.42 & & 0.67 & 0.16 \\
               & Teleoperated    & 100\%  &  25.86 &  23.09 & & 0.85 & 0.06 \\

        \bottomrule
    \end{tabular}
    \end{small}
    \caption{Results of the proposed robot FSM compared to the reference simulated teleoperated strategy.}
    \label{tab:exp_5}
\end{table}

In the final experiment, we compare our proposed FSM to a strategy where there is a simulated teleoperated goal robot (STR) present at all times, and which traces a path between the object and the goal while maintaining a constant distance between itself and the object. This strategy was previously proposed by \cite{Chen2015} to handle the presence of obstacles. The results in each environment are shown in \autoref{tab:exp_5} As expected, the STR method outperforms our proposed method in all environments in terms of path efficiency and mean trial duration, while the variance is also always lower in both of these metrics. This is because the STR is able to lead the goal around a pre-defined and efficient path around the obstacles and towards the goal. In contrast, our proposed FSM displays high variance and often worse average performance. The STR method achieves a higher success rate than our proposed FSM in the Middle environment (100\% vs. 76\%, respectively), as it does not struggle with the possibility of multiple paths between the object and the goal; the STR method is manually assigned one of the two paths. However, we note that the completion rate on the Reference and Corner environments of our proposed strategy is also at 100\% and that it reaches 94\% in the 2-Corner environment, suggesting its ability to handle environments where obstacles are present, and where the environment is not known prior to the object transportation task.

\section{Conclusion}\label{sec:conclusion}

In this paper, we augmented an occlusion-based cooperative object transportation strategy, from \cite{Chen2015}, such that a swarm of robots can collectively navigate objects around obstacles. The proposed strategy uses a large number of relatively small and easy-to-construct mobile robots to transport objects. Each robot is equipped with a simple FSM wherein the robots' behaviours are dictated entirely by visual cues. The strategy, as in \cite{Chen2015}, has robots push objects whenever the line-of-sight towards the goal is occluded by the object itself. Our proposed strategy adds an additional behaviour that enables the robots to form sub-goals; overcoming the limitation of the original occlusion-based strategy in environments where obstacles impede the line-of-sight between the object and the goal. Using a physics-based simulator, we demonstrate that our proposed strategy can complete the object transportation task in environments requiring the object to be navigated around at least one such obstacle or where multiple paths to the goal are possible. We also show that our strategy is effective for different convex and concave object shapes. 

The primary advantage of our approach is that it maintains the low-complexity nature of the original occlusion-based strategy. That is, the decentralised system functions using a number of relatively simple robots, without the need for any explicit communication with one another. Each robot executes an identical copy of a state machine algorithm, which does not require any explicit knowledge of the environment and operates purely with information within their local perceptual range. A benefit of our system is that it is robust to the complete failure or breakdown of one or more robots; as each robot's controller is identical, the remaining robots are still able to complete the object transportation task. We also demonstrate that the system scales well with more robots, and is not overly sensitive to a specific number of robots. 

However, in this study we did not examine how to extrapolate the optimal number of robots for any given object transportation setting; we are of the view that the optimal number of robots is a function of the size and mass of the object, as well as the size and complexity of the environment dictate the amount of force (robots) needed to push the object (as shown in Section \ref{sec:Different_Shape_Experiments}). For example, we would expect that an object with more mass would require more robots to push the object. In this study, we only experimented with the swarm density by altering the number of robots. As discussed by \cite{kuckling2024}, swarm density is also a function of the area of the environment. Due to the wide variety of possible configurations of the object transportation task, and the nuances of each aspect of the task, we do not believe we can prescribe a formula for the optimal number of robots in any given object transportation task within the current work. We also did not examine the influence of the object's mass on the completion or efficiency of the object transportation task using our proposed FSM. Further research is required in order to examine a possible heuristic for identifying the number of robots needed for the object transportation task in practical or commercial applications.

In future works, the strategy may still be improved. For instance, the use of random walking when the robots are initialised and looking for the object may be replaced by more explicit path-finding methods, in order to find the object more efficiently and more quickly, especially in more complex maze-like environments. Explicit path planning methods may also alleviate the low path efficiency in the Middle environment, an environment which suggests that the presence of multiple possible paths for the object to take may prove challenging for the current implementation of the FSM. Further work is needed to explore whether our proposed FSM remains effective when multiple directions lead towards the goal, especially in the presence of symmetric environments, where two paths may be equally `good' as one another. In the current implementation, no explicit mechanism is used to select a path when multiple chains of sub-goals are formed independently. A related issue observed in the experiments is the issue of synchronisation. Under the current implementation, two robots could transition into the sub-goal (S5) state at the exact same timestep; which may exacerbate the aforementioned two-paths issue. While we did not observe this phenomenon during our experiments, there may be scenarios not covered in the current work where the synchronisation between robots does affect task completion performance. Moreover, improvements to the placement of the robots along the object's perimeter may also be optimised further, such that the robots are more efficiently distributed and better placed to push the object such that excessive torque (and the resulting rotation) is not applied to the object when trying to move it across the environment. As was seen in Section \ref{sec:Different_Shape_Experiments}, the wall-following algorithm which distributes the robots around the correct sides of the object may not always move the robots to the correct position quickly enough if the object is already moving at a velocity similar to the robots' maximum speed; which may cause robots to block or inhibit the path of the object. One simple adjustment may be to reduce the speed at which the robots push the object, however, we believe that this would also come at a cost of increased task completion duration. Furthermore, implementing explicit strategies that prevent the object (or robots) from straying too close to obstacles is likely to reduce the likelihood of the object from being blocked or stuck up against a wall or other obstacle, as was observed in the experiments with the rectangular shape on the 2-Corners environment. Finally, all simulations were performed assuming ideal sensing conditions; noise in either the visual or proximity sensors in real-world applications would likely introduce inaccuracies in detecting (sub-)goals and obstacles, potentially reducing the overall path efficiency. Future work should investigate the impact of sensor noise and explore filtering or sensor fusion techniques to enhance the system's performance under more realistic conditions.

\backmatter

\bmhead{Acknowledgments}

We would like to thank Harmen de Weerd for his guidance throughout this project, as well as his support throughout the publication process.

\section*{Declarations}

\subsection{Funding}
The authors did not receive support from any organisation for the submitted work.

\subsection{Competing Interests}
The authors have no relevant competing interests or non-financial interests to disclose.

\subsection{Ethics Approval}
Not applicable.

\subsection{Consent to participate}
Not applicable.

\subsection{Consent for publication}
Not applicable.

\subsection{Availability of data and materials}
Not applicable.

\subsection{Code availability}
The code for this project is available at: 
\url{https://github.com/brenocq/object-transportation}

\subsection{Authors' contributions}
Both authors contributed to the conception and design of this study. The implementation of the code was performed by Breno Cunha Queiroz. The first draft of the manuscript was written by Daniel MacRae and both authors commented on previous versions of the manuscript. Both authors read and approved the final manuscript.

\begin{appendices}

\section{Details of the Proposed FSM Implementation}\label{sec:appendixA}

\subsection{Angles Between the Object and Goals}

\begin{figure}[ht]
    \centering
    \includegraphics[width=\linewidth]{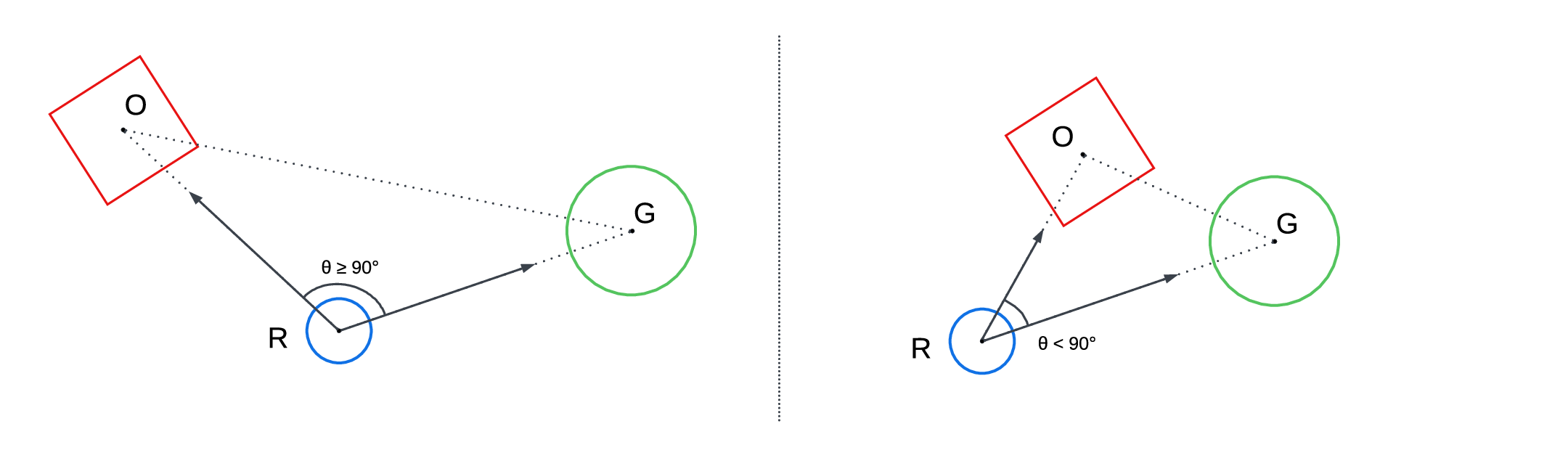}
    \caption{Diagram of the relative angles between the object and a goal from the point of view of a robot. The robot ($R$) is a blue circle, the goal ($G$) is a green circle, and the object ($O$) is a red square.}
    \label{fig:robot_angles}
\end{figure}

The transition between the S1 (Random Walk) state and S5 (Be a Goal) state includes a condition where the observed angle $\theta$ between the object and a goal, from the point of view of the robot, must have been at least 90\textdegree before the state transition into S5 is possible. This requirement is illustrated in Figure \ref{fig:robot_angles}. On the left side of this illustration is a case where this angle criterion is met, that is the angle is greater than 90\textdegree, and a transition into S5 would be allowed if the goal became occluded. On the right side is a case where the criterion is not met, and the robot would not be allowed to transition into state S5. 

This criterion was developed to prevent the robots from forming sub-goals in undesirable positions. For instance, in the right-hand side of Figure \ref{fig:robot_angles}, if the robot formed a sub-goal on (or around) its current position, it would cause the object to be pushed away from the main goal, as the sub-goal would not lie near the shortest path between the object's current position and the goal. On the left-hand side of Figure \ref{fig:robot_angles}, the extension of the path would not be very large.

\subsection{Determining When to Push the Object}

State S3 (Push Object) requires that the robot has a free space along the edge of the object in order to be able to push it. The algorithm to find free space for pushing the red object works by analyzing a camera image where the object and surrounding robots are represented by different colors. The robot detects a free space by scanning horizontally for any red pixels (indicating the object) that do not have blue pixels (indicating other robots) directly below them. If such a red pixel is found without a corresponding blue pixel underneath, this indicates an open space where the object can be pushed. If there are blue pixels underneath all of the red pixels, then there is no space from which the robot can push the object (and thus the robot moves to the S4 (Move Around Object) state).

\end{appendices}

\bibliography{sn-bibliography}% common bib file

@article{Alkilabi2017,
   author = {M H M Alkilabi and Aparajit Narayan and Elio Tuci},
   doi = {10.1007/s11721-017-0135-8},
   issn = {19353820},
   issue = {3-4},
   journal = {Swarm Intelligence},
   month = {12},
   pages = {185-209},
   publisher = {Springer New York LLC},
   title = {Cooperative object transport with a swarm of e-puck robots: robustness and scalability of evolved collective strategies},
   volume = {11},
   year = {2017},
}

@article{Bayindir2016,
   author = {Levent Bayindir},
   doi = {10.1016/J.NEUCOM.2015.05.116},
   issn = {0925-2312},
   journal = {Neurocomputing},
   month = {1},
   pages = {292-321},
   publisher = {Elsevier},
   title = {A review of swarm robotics tasks},
   volume = {172},
   year = {2016},
}

@article{Berman2011,
   author = {Spring Berman and Quentin Lindsey and Mahmut Selman Sakar and Vijay Kumar and Stephen C Pratt},
   doi = {10.1109/JPROC.2011.2111450},
   issue = {9},
   journal = {Proceedings of the IEEE},
   pages = {1470-1481},
   title = {Experimental Study and Modeling of Group Retrieval in Ants as an Approach to Collective Transport in Swarm Robotic Systems},
   volume = {99},
   year = {2011},
}

@article{Bernard2011,
   author = {Markus Bernard and Konstantin Kondak and Ivan Maza and Anibal Ollero},
   doi = {10.1002/rob.20401},
   issue = {6},
   journal = {Journal of Field Robotics},
   pages = {914-931},
   title = {Autonomous transportation and deployment with aerial robots for search and rescue missions},
   volume = {28},
   year = {2011},
}

@article{Brown1995,
   author = {R G Brown and J S Jennings},
   doi = {10.1109/IROS.1995.525941},
   journal = {Proceedings 1995 IEEE/RSJ International Conference on Intelligent Robots and Systems. Human Robot Interaction and Cooperative Robots},
   pages = {562-568},
   title = {A pusher/steerer model for strongly cooperative mobile robot manipulation},
   volume = {3},
   year = {1995},
}

@article{Carrillo-Zapata2020,
   author = {Daniel Carrillo-Zapata and Emma Milner and Julian Hird and Georgios Tzoumas and Paul J Vardanega and Mahesh Sooriyabandara and Manuel Giuliani and Alan F T Winfield and Sabine Hauert},
   doi = {10.3389/frobt.2020.00053},
   issn = {2296-9144},
   journal = {Frontiers in Robotics and AI},
   title = {Mutual Shaping in Swarm Robotics: User Studies in Fire and Rescue, Storage Organization, and Bridge Inspection},
   volume = {7},
   year = {2020},
}

@article{Chen2013,
   author = {Jianing Chen and Melvin Gauci and Roderich Groß},
   doi = {10.1109/ICRA.2013.6630674},
   isbn = {9781467356411},
   issn = {10504729},
   journal = {2013 IEEE International Conference on Robotics and Automation},
   pages = {863869},
   publisher = {IEEE Press},
   title = {A Strategy for Transporting Tall Objects with a Swarm of Miniature Mobile Robots},
   year = {2013},
}

@article{Chen2015,
   author = {Jianing Chen and Melvin Gauci and Wei Li and Andreas Kolling and Roderich Groß},
   doi = {10.1109/TRO.2015.2400731},
   issue = {2},
   journal = {IEEE Transactions on Robotics},
   pages = {307-321},
   title = {Occlusion-Based Cooperative Transport with a Swarm of Miniature Mobile Robots},
   volume = {31},
   year = {2015},
}

@article{Farivarnejad2021,
   author = {Hamed Farivarnejad and Amir Salimi Lafmejani and Spring Berman},
   doi = {10.1109/AERO50100.2021.9438133},
   journal = {2021 IEEE Aerospace Conference (50100)},
   pages = {1-9},
   title = {Fully Decentralized Controller for Multi-Robot Collective Transport in Space Applications},
   year = {2021},
}

@article{Farivarnejad2022,
   author = {Hamed Farivarnejad and Spring Berman},
   doi = {10.1146/annurev-control-042920-095844},
   issue = {1},
   journal = {Annual Review of Control, Robotics, and Autonomous Systems},
   pages = {205-219},
   title = {Multirobot Control Strategies for Collective Transport},
   volume = {5},
   year = {2022},
}

@article{Gelblum2016,
   author = {Aviram Gelblum and Itai Pinkoviezky and Ehud Fonio and Nir S Gov and Ofer Feinerman},
   doi = {10.1073/pnas.1611509113},
   issue = {51},
   journal = {Proceedings of the National Academy of Sciences},
   pages = {14615-14620},
   title = {Emergent oscillations assist obstacle negotiation during ant cooperative transport},
   volume = {113},
   year = {2016},
}

@article{Gros2009,
   author = {Roderich Groß and Marco Dorigo},
   doi = {10.1504/IJBIC.2009.022770},
   journal = {International Journal of Bio-Inspired Computation},
   month = {11},
   title = {Towards Group Transport by Swarms of Robots},
   volume = {1},
   pages = {1-13},
   year = {2009},
}

@article{Guo2017,
   author = {Shihui Guo and Meili Wang and Gabriel Notman and Jian Chang and Jianjun Zhang and Minghong Liao},
   doi = {10.1002/cav.1779},
   issue = {3-4},
   journal = {Computer Animation and Virtual Worlds},
   pages = {e1779},
   title = {Simulating collective transport of virtual ants},
   volume = {28},
   year = {2017},
}

@article{Habibi2016,
   author = {Golnaz Habibi and William Xie and Mathew Jellins and James McLurkin},
   doi = {10.1007/978-4-431-55879-8\_11/COVER},
   isbn = {9784431558774},
   issn = {1610742X},
   journal = {Springer Tracts in Advanced Robotics},
   pages = {151-164},
   publisher = {Springer Verlag},
   title = {Distributed path planning for collective transport using homogeneous multi-robot systems},
   volume = {112},
   year = {2016},
}

@article{Hu2011,
   author = {Wenqi Hu and Kelly S. Ishii and Aaron T. Ohta},
   doi = {10.1063/1.3631662},
   issn = {0003-6951},
   issue = {9},
   journal = {Applied Physics Letters},
   month = {8},
   pages = {094103},
   title = {Micro-assembly using optically controlled bubble microrobots},
   volume = {99},
   year = {2011},
}

@inbook{Huntsberger2000,
    author = {Terry Huntsberger  and Guillermo Rodriguez  and Paul S. Schenker },
    title = {Robotics Challenges for Robotic and Human Mars Exploration},
    booktitle = {Robotics 2000},
    chapter = {},
    pages = {340-346},
    doi = {10.1061/40476(299)45},
    editor = {WC Stone},
    year = {2012}

}

@article{Jurt2022,
   author = {Marius Jurt and Emma Milner and Mahesh Sooriyabandara and Sabine Hauert},
   doi = {10.1007/s10015-022-00730-5},
   issn = {16147456},
   issue = {2},
   journal = {Artificial Life and Robotics},
   month = {5},
   pages = {365-372},
   publisher = {Springer Japan},
   title = {Collective transport of arbitrarily shaped objects using robot swarms},
   volume = {27},
   year = {2022},
}

@article{Kosuge1996,
   author = {K Kosuge and T Oosumi},
   doi = {10.1109/IROS.1996.570694},
   journal = {Proceedings of IEEE/RSJ International Conference on Intelligent Robots and Systems. IROS '96},
   pages = {318-323},
   title = {Decentralized control of multiple robots handling an object},
   volume = {1},
   year = {1996},
}

@article{Kube2000,
   author = {C Ronald Kube and Eric Bonabeau},
   journal = {Robotics and Autonomous Systems},
   pages = {85-101},
   title = {Cooperative transport by ants and robots},
   volume = {30},
   year = {2000},
    doi = {https://doi.org/10.1016/S0921-8890(99)00066-4},
}

@article{Kube1997,
   author = {C. Ronald Kube and Hong Zhang},
   doi = {10.1023/A:1008859119831},
   issn = {09295593},
   issue = {1},
   journal = {Autonomous Robots},
   pages = {53-72},
   publisher = {Kluwer Academic Publishers},
   title = {Task Modelling in Collective Robotics},
   volume = {4},
   year = {1997},
}

@article{Ligot2020,
   author = {Antoine Ligot and Jonas Kuckling and Darko Bozhinoski and Mauro Birattari},
   doi = {10.7717/PEERJ-CS.314/SUPP-4},
   issn = {23765992},
   journal = {PeerJ Computer Science},
   month = {11},
   pages = {1-27},
   publisher = {PeerJ Inc.},
   title = {Automatic modular design of robot swarms using behavior trees as a control architecture},
   volume = {6},
   year = {2020},
}

@article{Michael2011,
   author = {Nathan Michael and Jonathan Fink and Vijay Kumar},
   doi = {10.1007/s10514-010-9205-0},
   issn = {1573-7527},
   issue = {1},
   journal = {Autonomous Robots},
   pages = {73-86},
   title = {Cooperative manipulation and transportation with aerial robots},
   volume = {30},
   year = {2011},
}

@article{Neumann2014,
   author = {Michael Neumann and Matthew Chin and C Kitts},
   journal = {Lecture Notes in Engineering and Computer Science},
   pages = {364-369},
   title = {Object Manipulation through Explicit Force Control Using Cooperative Mobile Multi-Robot Systems},
   volume = {1},
   year = {2014},
}

@article{Nouyan2008,
   author = {Shervin Nouyan and Alexandre Campo and Marco Dorigo},
   doi = {10.1007/s11721-007-0009-6},
   issn = {1935-3820},
   issue = {1},
   journal = {Swarm Intelligence},
   pages = {1-23},
   title = {Path formation in a robot swarm},
   volume = {2},
   year = {2008},
}

@article{Parker2006,
   author = {Chris A C Parker and Hong Zhang},
   doi = {10.1177/105971230601400101},
   issue = {1},
   journal = {Adaptive Behavior},
   pages = {5-19},
   title = {Collective Robotic Site Preparation},
   volume = {14},
   year = {2006},
}

@article{Rahman2017,
   author = {M Arifur Rahman and Julian Cheng and Zhidong Wang and Aaron T Ohta},
   doi = {10.1038/s41598-017-03525-y},
   issn = {2045-2322},
   issue = {1},
   journal = {Scientific Reports},
   pages = {3278},
   title = {Cooperative Micromanipulation Using the Independent Actuation of Fifty Microrobots in Parallel},
   volume = {7},
   year = {2017},
}

@article{Rauniyar2021,
   author = {Aditya Rauniyar and Hem Chandra Upreti and Aman Mishra and Prabhu Sethuramalingam},
   doi = {10.1007/s10846-021-01359-5},
   issn = {1573-0409},
   issue = {1},
   journal = {Journal of Intelligent \& Robotic Systems},
   pages = {3},
   title = {MeWBots: Mecanum-Wheeled Robots for Collaborative Manipulation in an Obstacle-Clustered Environment Without Communication},
   volume = {102},
   year = {2021},
}

@article{Roodbergen2009,
   author = {Kees Jan Roodbergen and Iris F A Vis},
   doi = {10.1016/j.ejor.2008.01.038},
   issn = {0377-2217},
   issue = {2},
   journal = {European Journal of Operational Research},
   pages = {343-362},
   title = {A survey of literature on automated storage and retrieval systems},
   volume = {194},
   year = {2009},
}

@article{Shahrokhi2016,
   author = {Shiva Shahrokhi and Aaron T Becker},
   doi = {10.1109/COASE.2016.7743453},
   journal = {2016 IEEE International Conference on Automation Science and Engineering (CASE)},
   pages = {561-566},
   title = {Object manipulation and position control using a swarm with global inputs},
   year = {2016},
}

@article{Sugie1995,
   author = {H Sugie and Y Inagaki and S Ono and H Aisu and T Unemi},
   doi = {10.1109/ROBOT.1995.525583},
   journal = {Proceedings of 1995 IEEE International Conference on Robotics and Automation},
   pages = {2181-2186},
   title = {Placing objects with multiple mobile robots-mutual help using intention inference},
   volume = {2},
   year = {1995},
}

@article{Takeda2002,
   author = {Hiroki Takeda and Yasuhisa Hirata and Zhi-Dong Wang and Kazuhiro Kosuge},
   doi = {10.1007/978-4-431-65941-9\_16},
   journal = {Distributed Autonomous Robotic Systems 5},
   pages = {155-164},
   publisher = {Springer, Tokyo},
   title = {Collision Avoidance Algorithm for Two Tracked Mobile Robots Transporting a Single Object in Coordination Based on Function Allocation Concept},
   year = {2002},
}

@article{Tuci2018,
   author = {Elio Tuci and Muhanad Alkilabi and Otar Akanyeti},
   doi = {10.3389/frobt.2018.00059},
   journal = {Frontier in Robotics and AI},
   month = {11},
   title = {Cooperative Object Transport in Multi-Robot Systems: A Review of the State-of-the-Art},
   volume = {5},
    pages = {59},
    ISSN={2296-9144},  
   year = {2018},
}

@article{Wang2016,
   author = {Zijian Wang and Mac Schwager},
   doi = {10.1109/ICRA.2016.7487163},
   journal = {2016 IEEE International Conference on Robotics and Automation (ICRA)},
   pages = {427-432},
   title = {Kinematic multi-robot manipulation with no communication using force feedback},
   year = {2016},
}

@article{Wang2006,
   author = {Ying Wang and C W de Silva},
   doi = {10.1109/CEC.2006.1688694},
   journal = {2006 IEEE International Conference on Evolutionary Computation},
   pages = {3050-3056},
   title = {Cooperative Transportation by Multiple Robots with Machine Learning},
   year = {2006},
}

@article{Wilson2014,
   author = {Sean Wilson and Theodore P Pavlic and Ganesh P Kumar and Aurélie Buffin and Stephen C Pratt and Spring Berman},
   doi = {10.1007/s11721-014-0100-8},
   issn = {1935-3820},
   issue = {4},
   journal = {Swarm Intelligence},
   pages = {303-327},
   title = {Design of ant-inspired stochastic control policies for collective transport by robotic swarms},
   volume = {8},
   year = {2014},
}

@article{Woern2006,
   author = {Heinz Woern and Marc Szymanski and Joerg Seyfried},
   doi = {10.1109/ROMAN.2006.314376},
   journal = {The 15th IEEE International Symposium on Robot and Human Interactive Communication},
   pages = {492-496},
   title = {The I-SWARM project},
   year = {2006},
}

@article{Sudsang_caging,
  author={Sudsang, A. and Ponce, J.},
  journal={2000 {IEEE} International Conference on Robotics and Automation}, 
  title={A new approach to motion planning for disc-shaped robots manipulating a polygonal object in the plane}, 
  year={2000},
  volume={2},
  number={},
  pages={1068-1075 vol.2},
  doi={10.1109/ROBOT.2000.844741}
}

@article{Wang_caging,
  author={Zhidong Wang and Hirata, Y. and Kosuge, K.},
  journal={Proceedings 2003 IEEE/RSJ International Conference on Intelligent Robots and Systems}, 
  title={Control multiple mobile robots for object caging and manipulation}, 
  year={2003},
  volume={2},
  number={},
  pages={1751-1756 vol.2},
  doi={10.1109/IROS.2003.1248897}
}

@article{Nouyan2009,
  author={Nouyan, Shervin and Gross, Roderich and Bonani, Michael and Mondada, Francesco and Dorigo, Marco},
  journal={IEEE Transactions on Evolutionary Computation}, 
  title={Teamwork in Self-Organized Robot Colonies}, 
  year={2009},
  volume={13},
  number={4},
  pages={695-711},
  doi={10.1109/TEVC.2008.2011746}
}

@article{Sugawara_pheromone,
  author={Sugawara, K. and Kazama, T. and Watanabe, T.},
  journal={2004 IEEE/RSJ International Conference on Intelligent Robots and Systems (IROS)}, 
  title={Foraging behavior of interacting robots with virtual pheromone}, 
  year={2004},
  volume={3},
  number={},
  pages={3074-3079 vol.3},
  doi={10.1109/IROS.2004.1389878}
}

@article{Campo2010ArtificialPF,
  title={Artificial pheromone for path selection by a foraging swarm of robots},
  author={Alexandre Campo and {\'A}lvaro Guti{\'e}rrez and Shervin Nouyan and Carlo Pinciroli and Valentin Longchamp and Simon Garnier and Marco Dorigo},
  journal={Biological Cybernetics},
  year={2010},
  volume={103},
  pages={339-352},
  doi = {https://doi.org/10.1007/s00422-010-0402-x}
}

@misc{Atta2020,
  author = {Breno Cunha Queiroz},
  title = {{Atta: A Large-Scale Multi-Robot Simulator}},
  howpublished = {\url{https://github.com/brenocq/atta}},
  year = {2020},
  note = {Accessed on 2nd August 2023},
}

@article{Zhang2020,
  author       = {Lin Zhang and
                  Hao Xiong and
                  Ou Ma and
                  Zhaokui Wang},
  title        = {Multi-robot Cooperative Object Transportation using Decentralized
                  Deep Reinforcement Learning},
  journal      = {CoRR},
  volume       = {abs/2007.09243},
  year         = {2020},
  url          = {https://arxiv.org/abs/2007.09243},
  eprinttype    = {arXiv},
  eprint       = {2007.09243},
  bibsource    = {dblp computer science bibliography, https://dblp.org}
}

@Inbook{Alkilabi2018,
    author="M H M Alkilabi
    and Narayan, Aparajit
    and Lu, Chuan
    and Tuci, Elio",
    editor="Gro{\ss}, Roderich
    and Kolling, Andreas
    and Berman, Spring
    and Frazzoli, Emilio
    and Martinoli, Alcherio
    and Matsuno, Fumitoshi
    and Gauci, Melvin",
    title="Evolving Group Transport Strategies for e-Puck Robots: Moving Objects Towards a Target Area",
    bookTitle="Distributed Autonomous Robotic Systems: The 13th International Symposium",
    year="2018",
    publisher="Springer International Publishing",
    address="Cham",
    pages="503--516",
    isbn="978-3-319-73008-0",
    doi="10.1007/978-3-319-73008-0_35",
}

@inbook{Ebel2021,
  author={Ebel, Henrik and Eberhard, Peter},
  booktitle={2021 International Symposium on Multi-Robot and Multi-Agent Systems (MRS)}, 
  title={Non-Prehensile Cooperative Object Transportation with Omnidirectional Mobile Robots: Organization, Control, Simulation, and Experimentation}, 
  year={2021},
  volume={},
  number={},
  pages={1-10},
  doi={10.1109/MRS50823.2021.9620541}}

@misc{kuckling2024,
      title={Do We Run Large-scale Multi-Robot Systems on the Edge? More Evidence for Two-Phase Performance in System Size Scaling}, 
      author={Jonas Kuckling and Robin Luckey and Viktor Avrutin and Andrew Vardy and Andreagiovanni Reina and Heiko Hamann},
      year={2024},
      eprint={2310.11843},
      archivePrefix={arXiv},
      primaryClass={cs.RO},
      url={https://arxiv.org/abs/2310.11843}, 
}

\end{document}